\documentclass{article}

    \usepackage[final]{neurips_2024}

\usepackage[utf8]{inputenc} 
\usepackage[T1]{fontenc}    
\usepackage{hyperref}       
\usepackage{url}            
\usepackage{booktabs}       
\usepackage{amsfonts}       
\usepackage{nicefrac}       
\usepackage{microtype}      
\usepackage{xcolor}         

\newcommand{\lyx}[1]{\textcolor{black}{#1}}
\newcommand{\mname}{SeTAR }
\newcommand{\mnamec}{SeTAR}
\usepackage[skip=0pt]{caption}

\usepackage{amsmath}
\usepackage{amsfonts}
\usepackage{algorithm}
\usepackage{algorithmic}
\usepackage[algo2e]{algorithm2e}
\usepackage{wrapfig}
\usepackage{graphicx}
\usepackage{multirow}
\usepackage{subfigure}
\usepackage{listings}
\usepackage{wrapfig}
\usepackage{caption}

\title{SeTAR: Out-of-Distribution Detection with \\ Selective Low-Rank Approximation}

\author{%
Yixia Li$^{1}$\thanks{Equal Contribution.}\;, \
Boya Xiong$^{2}$\footnotemark[1]\;, \
Guanhua Chen$^{1}$\thanks{Corresponding Authors.}\;, \
Yun Chen$^{3}$\footnotemark[2]  \\
$^1$Southern University of Science and Technology \\
$^2$Shanghai University of Finance and Economics  \\
$^3$MoE Key Laboratory of Interdisciplinary Research of Computation and Economics, \\
Shanghai University of Finance and Economics \\
\texttt{liyixia@me.com,  xiongboya@163.sufe.edu.cn} \\
\texttt{chengh3@sustech.edu.cn,  yunchen@sufe.edu.cn}
}

\begin{document}

\maketitle

\begin{abstract}

Out-of-distribution (OOD) detection is crucial for the safe deployment of neural networks. Existing CLIP-based approaches perform OOD detection by devising novel scoring functions or sophisticated fine-tuning methods. In this work, we propose SeTAR, a novel, training-free OOD detection method that leverages selective low-rank approximation of weight matrices in vision-language and vision-only models. SeTAR enhances OOD detection via post-hoc modification of the model's weight matrices using a simple greedy search algorithm. Based on SeTAR, we further propose SeTAR+FT, a fine-tuning extension optimizing model performance for OOD detection tasks. Extensive evaluations on ImageNet1K and Pascal-VOC benchmarks show SeTAR's superior performance, reducing the \lyx{relatively} false positive rate by up to 18.95\% and 36.80\% compared to zero-shot and fine-tuning baselines. Ablation studies further validate SeTAR's effectiveness, robustness, and generalizability across different model backbones. Our work offers a scalable, efficient solution for OOD detection, setting a new state-of-the-art in this area.

\end{abstract}

\section{Introduction}
\label{intro}

The task of out-of-distribution (OOD) detection \citep{hendrycks2016baseline,ming2022delving} aims to identify whether input data comes from an unknown distribution. It has garnered significant attention in the machine learning community recently \citep{hendrycks-etal-2020-pretrained,xu-etal-2021-unsupervised,miyai2023locoop}.
While machine learning models are trained with supervised in-distribution (ID) data, they often struggle to generalize to OOD data encountered in real-world applications \citep{emmott2016metaanalysis} like autonomous vehicles and healthcare.
These OOD samples pose challenges as they are not represented in the training data. Consequently, OOD detection plays a crucial role in developing reliable and trustworthy machine-learning models suitable for real-world deployment \citep{bai2023feed}. It allows models to filter out and reject these awkward inputs effectively, and enables the use of curated and labeled OOD samples to further train for a more robust model in the wild.

Previous research has primarily focused on detecting OOD instances in either visual \citep{devries2018learning,liang2017enhancing,hendrycks2019scaling} or textual data \citep{hu2021uncertainty,zheng2020ooddialog,zhou-etal-2021-contrastive}.
Recently, significant progress has been made in multimodal tasks like multimodal retrieval \citep{li2023blip2,caesar2018coco} and image classification \citep{yu2022coca}, thanks to vision-and-language pretrained (VLP) models like CLIP \citep{radford2021learning}.
More recent studies have explored OOD detection with CLIP, grouped into zero-shot methods \citep{fort2021exploring,ming2022delving,miyai2023zero} and finetuning-based methods \citep{Ming_2023,tao2023non,miyai2023locoop}. However, the zero-shot methods suffer from suboptimal performance due to potential domain gaps with ID downstream data. On the other hand, finetuning-based methods carry the risk of deconstructing the intricate representations learned by CLIP which requires a meticulously designed training strategy.
\lyx{Sparsification-based approaches \citep{sun2021react,djurisic2023extremely} have demonstrated potential in OOD detection within CNNs, leveraging the assumption that ID and OOD samples produce distinct activation patterns. Nevertheless, their effectiveness diminishes in large-scale pre-trained models such as CLIP, where activation differences become more subtle, thereby limiting their applicability primarily to models fine-tuned on downstream ID-domain datasets.}

In this work, we propose SeTAR, a training-free and effective OOD detection method by selective low-rank approximations.
Low-rank approximation is to approximate a given matrix by finding a lower-rank matrix that closely resembles the original matrix. Previous research has demonstrated that using low-rank approximation matrices can achieve comparable performance to full parameters in various scenarios, as observed in tasks such as large language model (LLM) fine-tuning~\citep{hu2022lora} and model pruning~\citep{Hajimolahoseini2021CompressingPL}. These approaches typically preserve the same rank across different low-rank approximation matrices.
In our work, we demonstrate that it is possible to significantly enhance the performance of OOD detection by selectively manipulating the weight matrices in the CLIP model, including the choice of weight matrices and the ratio of singular vectors to be reduced. Specifically, we propose a simple top-to-bottom and image-to-text greedy search algorithm to manipulate $\mathrm{W_{\text{up}}}$ in the CLIP model. Our method applies to various model backbones and does not require any additional training or new parameters. Building upon \mname, we further demonstrate its effectiveness for fine-tuning initialization, referred to as \mnamec+FT.

We conduct extensive evaluations and achieve state-of-the-art performance on common OOD detection benchmarks for CLIP, including the ImageNet1K and Pascal-VOC benchmarks. Compared to vanilla MCM and GL-MCM, \mname with the CLIP backbone reduces \lyx{relatively} FPR95 by 9.5\% and 12.0\% on average across two benchmarks, respectively.
When further integrate fine-tuning into \mnamec, \mnamec+FT outperforms the state-of-the-art fine-tuning baselines LoCoOp~\citep{miyai2023locoop} and LoRA~\citep{hu2022lora}. Moreover, we perform a comprehensive ablation study and analysis to verify and understand \mnamec. In summary, our key results and contributions:

\begin{enumerate}
    \item We propose SeTAR, a simple yet effective OOD detection method based on selective low-rank approximation. It is training-free as it only performs post-hoc modification to weight matrices. \mname applies to a variety of scoring functions and model backbones. It can be readily integrated with existing zero-shot OOD detection methods.
    \item We further extend \mname to \mnamec+FT, which demonstrates the effectiveness of \mname in improving the performance of finetuning-based OOD detection methods and achieving new state-of-the-art results.
    \item We extensively evaluate \mname and \mnamec+FT across a diverse set of OOD detection tasks. It consistently outperforms baseline methods and establishes new state-of-the-art results on CLIP-based OOD detection benchmarks. On ImageNet1K, \mname achieves an AUROC of 91.32\% with CLIP backbone and GL-MCM score. The score further increases to 92.31\% when combined with the finetuning-based detection method.
    \item We perform comprehensive ablation studies and empirical analyses to verify and understand \mnamec. We hope that this work will shed light on future explorations on in-depth understanding of the \mname method.\footnote{\ \ Code are available at \ \url{https://github.com/X1AOX1A/SeTAR}.}
\end{enumerate}

\begin{figure*}[t]
    \centering
    \includegraphics[width=0.8\textwidth]{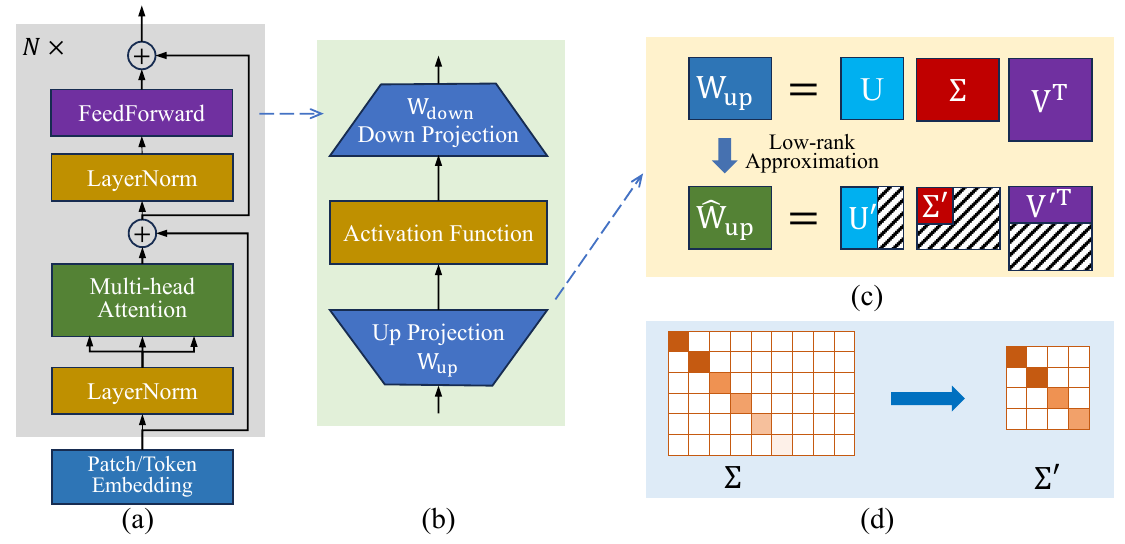}    
    \caption{The overview of \mnamec. (a) The structure of the CLIP image and text encoder. (b) The details of the feed-forward sublayer. (c) For each encoder layer, we replace the $\mathrm{W_{\text{up}}}$ weight matrix with its low-rank approximation $\mathrm{\widehat{W}_{\text{up}}}$. (d) The illustration of $\Sigma$ before and after low-rank approximation. More details are in Section~\ref{sec:lowrapx}.}
    \label{fig:overview}
    \vspace{-10pt}
\end{figure*}

\section{Preliminaries}
\paragraph{CLIP Architecture}

The CLIP model \citep{radford2021learning} comprises an image encoder $E^v(\cdot)$ and a text encoder $E^t(\cdot)$, aligned via contrastive learning on web-scale image-text pairs. We focus on CLIP-ViT, where the image encoder is a Vision Transformer (ViT). Each ViT layer includes a multihead self-attention sublayer and a feed-forward sublayer. In the self-attention module, the hidden state is projected into different spaces using learnable parameters $\mathrm{W_q, W_k, W_v}$. The outputs are concatenated and projected back with another linear matrix $\mathrm{W_o}$. The feed-forward module projects the hidden state into a wider space using $\mathrm{W_{\text{up}}}$ and then back with $\mathrm{W_{\text{down}}}$ after a non-linear activation (Figure~\ref{fig:overview}).
Given the similarity between the image and text encoder layers, we adopt consistent notations for the linear matrices in both. Each encoder also includes a linear projector $\mathrm{W_{p}}$ to map their representations into a shared space for contrastive learning.

\paragraph{Zero-shot OOD Detection with CLIP}
Zero-shot OOD detection aims to separate ID and OOD data without an ID training dataset. Given the CLIP, the ID classes are defined by the classification task of interest, which differs from the classes used in CLIP pretraining.
Accordingly, OOD is defined as classes not belonging to any ID class, making the OOD detector a binary classifier. MCM~\citep{ming2022delving} and GL-MLM~\citep{miyai2023zero} are two zero-shot CLIP-based OOD detection methods.
Formally, let $\mathbf{x}$ be the test image and $\mathcal{T}_{\text{in}}=\{\mathbf{y}_c\}_{c=1}^K$ be the set of text prompts containing $M$ ID class labels (e.g., "a photo of a [CLASS]"). The image is segmented into $l$ image patches $\mathbf{x}=(x_1,...,x_l)$. Following CLIP, we add $\text{[cls]}$ before the image patches and use the output of $\text{[cls]}$ from the visual projector $\mathrm{W_{p}}$ as the global image feature ($h^v \in \mathbb{R}^d$).
The outputs of other patches are projected by the visual projector as the local image features ($\mathbf{p}^v=(p^v_1,...,p^v_l)\in \mathbb{R}^{l\times d}$). For the text prompt $\mathbf{y}_c \in \mathcal{T}_{\text{in}}$, we add an additional $\text{[eos]}$ after the text tokens and use the output feature of $\text{[eos]}$ from the textual projector $\mathrm{W_{p}}$ as the concept feature of ID class $c$ ($h^t_c \in \mathbb{R}^d$).

The label-wise image-concept matching (IWIC) score measures how well a test image $\mathbf{x}$ aligns with a concept $\mathbf{y}_c$, using either global or local features. The global IWIC score $s^G_c\left(\cdot\right)$ is the cosine similarity between the global image feature $h^v$ and the concept feature $h^t_c$: $s^G_c\left(\mathbf{x}\right)=\mathrm{cos\_sim}(h^v, h^t_c)$. The local IWIC score $s^L_c(\cdot)$ is the max-pooled cosine similarity between image patch features $p_i^v$ and the concept feature $h^t_c$: $s^L_c(\mathbf{x}) =\max _{i} \mathrm{cos\_sim}(p_i^v, h^t_c)$. The MCM and GL-MCM scores are defined as:

\vspace{-10pt}
\begin{equation}
\label{eq:mcm_cal}
    S_{\mathrm{MCM}}\left(\mathbf{x}\right) =\max _c \frac{e^{s^G_c\left(\mathbf{x}\right) / \tau}}{\sum_{c=1}^K e^{s^G_c\left(\mathbf{x}\right) / \tau}},
\end{equation}
\begin{equation}
\label{eq:glmcm_cal}
    S_{\mathrm{GL-MCM}}\left(\mathbf{x}\right)=S_{\mathrm{MCM}}\left(\mathbf{x}\right)+\max _c \frac{e^{s_c^L\left(\mathbf{x}\right) / \tau^\prime}}{\sum_{c=1}^K e^{s_c^L\left(\mathbf{x}\right) / \tau^\prime}},
\end{equation}
where $\tau$ and $\tau^\prime$ are the temperature hyperparameters. MCM only uses global image features, while GL-MCM additionally considers local image features. For ID data, both MCM and GL-MCM scores will be matched to one of the concept features with a high score; and vice versa. As a result, our OOD detection function can be formulated as:
\begin{equation}
    G\left(\mathbf{x}\right)=\left\{\begin{array}{ll}
1 & S\left(\mathbf{x}\right) \geq \lambda \\
0 & S\left(\mathbf{x}\right)<\lambda
\end{array},\right.
\label{eqn:detec}
\end{equation}
where $S\left(\mathbf{x}\right)$ is either the MCM or GL-MCM score, $\lambda$ is the threshold value. By convention, $G(\mathbf{x})=1$ represents the ID class and $G(\mathbf{x})=0$ indicates the OOD class. The $\lambda$ is chosen so that a high fraction of ID data (e.g., 95\%) is above the threshold.
We follow previous work~\citep{miyai2023locoop} to use either MCM or GL-MCM score for OOD detection in this work.

\section{Method}
We introduce \mnamec, a training-free and effective technique for improving OOD detection performance (see Figure~\ref{fig:overview}). Our key idea is to perform post-hoc modification to CLIP weight matrices by selectively replacing them with their low-rank approximations. It is complementary to existing CLIP-based zero-shot OOD detection methods and could be further extended to finetuning-based methods, which we term as \mnamec+FT.

\subsection{OOD Detection with Selective Low-Rank Approximation}
\paragraph{Low-Rank Approximation} \label{sec:lowrapx}
Given a linear matrix $W \in \mathbb{R}^{m \times n}$, its Singular Value Decomposition (SVD) is denoted as $W=U \Sigma V^{\top}$, where $U=\left[u_1, u_2, \cdots, u_m\right] \in \mathbb{R}^{m \times m}$, $V=\left[v_1, v_2, \cdots, v_n\right] \in \mathbb{R}^{n \times n}$,
and $\Sigma \in \mathbb{R}^{m \times n}$ is a matrix whose entries are all zero except for the singular values of $W$.
These singular values appear in decreasing order on the diagonal (i.e. $\sigma_i^{\downarrow}(W)$). The SVD of $W$ can be reformulated as in Equation~\ref{eq:svd-decomp}.
Given a hyperparameter $r \in \mathbb{N}^+$, a rank-$r$ approximation of $W$ is matrix $\widehat{W}$ that minimizes $\|W-\widehat{W}\|_2$ and satisfies $\text{rank}(\widehat{W})\leq r$.
The optimal solution of this problem $\widehat{W}$ is provided by Eckart–Young–Mirsky theorem \citep{LowrankApproximation2024} using Singular Value Decomposition (see Equation~\ref{eq:lr-apox}).

\vspace{-15pt}
\begin{align}
W&=\sum_{i=1}^{\min(m, n)} \sigma_i^{\downarrow}(W) u_i v_i^{\top},  \label{eq:svd-decomp}\\
\widehat{W}&=\sum_{i=1}^r \sigma_i^{\downarrow}(W) u_i v_i^{\top}.  \label{eq:lr-apox}
\end{align}
\vspace{-10pt}

In this work, we will use the term \textbf{minor singular components} to refer to entries in the SVD corresponding to small singular values. These components are removed in low-rank approximation. The term of \textbf{principle singular components} is used to refer to entries in the SVD corresponding to large singular values. These components are kept in a low-rank approximation of the matrix.

\paragraph{OOD Detection with Selective Low-Rank Approximation}

\lyx{SVD-based weight pruning, particularly in noise-prone layers, can substantially reduce a network’s sensitivity to minor perturbations, leading to enhanced stability and robustness \citep{yao2024enhancingincontextlearningperformance}. This stability is crucial for OOD detection, as it ensures the model's reliable performance across a wide range of inputs.}
Building on this, we propose a method to improve OOD detection by selectively applying low-rank approximation to weight matrices. By decomposing a weight matrix $W$ into its singular values and vectors, we can identify and retain the principle singular components that significantly contribute to the model's performance. This approach ensures that the essential features of $W$ are preserved while discarding the less critical minor singular components.
Given a weight matrix $W$ in CLIP (e.g., $\mathrm{W_{\text{up}}}$ or $\mathrm{W_{k}}$), we replace the matrix with its low-rank approximation part $\widehat{W}$ as described in Equation~\ref{eq:lr-apox} (see Figure~\ref{fig:overview}). Given the rank reduction ratio $\Theta$, the rank of $\widehat{W}$ is determined by $r(\widehat{W}) = \text{round}((1 - \Theta) \cdot r(W))$.
This selective low-rank approximation leverages the compact representation provided by SVD to enhance the model's ability to detect OOD instances effectively without requiring additional training. We demonstrate our method's ability to improve OOD detection (Table~\ref{tab:main}) while maintaining ID classification performance (Table~\ref{tab:classfiy_acc}) in Section~\ref{sec:train-free} and Section~\ref{sec:analyses}.










\paragraph{HyperParameter Search Algorithm} \label{sec:hyp_search}
Due to the presence of many weight matrices in CLIP, each consisting of hundreds of singular values, conducting a complete search over all combinations of low-rank approximation weight matrices is impractical. Therefore, we propose a greedy search algorithm to determine the rank reduction ratio for each weight matrix.
Among all linear weight matrices in each encoder layer, we focus on $\mathrm{W_{\text{up}}}$ as it is most effective according to our preliminary experiment. For simplicity, we assume both image and text encoders have $N$ encoder layers.
As shown in Algorithm~\ref{alg:greedy_search}, we search by first enumerating all $N$ vision encoder layers sequentially from top to bottom and then all $N$ text encoder layers in the same way. This search order is concisely denoted as searching from $2N$ to the first layer in CLIP.  We compare different search algorithms in Section~\ref{sec:dif-search}.
The rank reduction ratio for each layer is the objective in \mname which is searched among the candidate list $\mathbf{\Theta}=\{\Theta_0, \Theta_1, \cdots, \Theta_J\}$ according to the loss on the validation set.
We employ the LoCoOp loss (Equation~\ref{eq:locoop_loss}) proposed in \citep{miyai2023locoop} as our loss function. This loss requires only ID images. It contains an ID loss for ID image classification and an OOD loss to push away pseudo OOD features from the ID class text embeddings where the pseudo OOD features are from ID-irrelevant nuisances (Equation~\ref{eq:ood_region}) (e.g., backgrounds) in CLIP’s local features. We refer the readers to \cite{miyai2023locoop} or Appendix~\ref{sec:locoop_loss} for more details.
\captionsetup[algorithm]{font=footnotesize}
\SetKwComment{Comment}{// }{}
\newcommand{\lIfElse}[3]{\lIf{#1}{#2 \textbf{else}~#3}}
\begin{wraptable}{r}{0.5\textwidth}
\centering
\vspace{-20pt}
\resizebox{0.5\textwidth}{!}{
  \begin{minipage}{0.5\textwidth}
    \begin{algorithm}[H]
    \footnotesize
    \caption{The hyperparameter search in SeTAR.}
    \label{alg:greedy_search}
    \KwData{Valid set $D$.}
    \KwIn{Layer length $2N$, rank reduction ratio candidates $\mathbf{\Theta}$ with length $J$, loss $\mathcal{L}_0$
 on $D$ WITHOUT SeTAR.}
    \KwResult{Rank reduction ratio list $\mathbf{T}^*$ with length $2N$.}
    $\mathcal{L}^* = \mathcal{L}_0$\; \Comment*[r]{Current best loss}
    \For{\textrm{LayerNum} $l \leftarrow 2N$ to $1$}{
        $\widehat{\mathrm{W}}^* = \mathrm{W^{l}_{\text{up}}}$\;
        $T^*[l] = 0$\;
        
        \For{\textrm{counter} $j \leftarrow 1$ to $J$}{

            $r = \text{round}((1 - \Theta[j]) \cdot \text{rank}(\mathrm{W^{l}_{\text{up}}}))$\;

            $\widehat{\mathrm{W}} = \sum_{i=1}^r \sigma_i^{\downarrow} u_i v_i^{\top}$\;

            Calcluate loss $\mathcal{L}_{j}^l$ on $D$ by replacing $\mathrm{W^{l}_{\text{up}}}$ with $\widehat{\mathrm{W}}$\;

            \If{$\mathcal{L}_{j}^l < \mathcal{L}^*$}{
                $\widehat{\mathrm{W}}^* = \widehat{\mathrm{W}}$;
                $T^*[l] = \Theta[j]$;
                $\mathcal{L}^* = \mathcal{L}_{j}^l$;
            }
        }
        $\mathrm{W^{l}_{\text{up}}} := \widehat{\mathrm{W}}^*$\;
    }
    \Return $T^*$\;
    \end{algorithm}
  \end{minipage}
}
\vspace{-25pt}
\end{wraptable}
For $\Theta_j \in \mathbf{\Theta}$, we remove $\Theta_j$ (in percent) singular values along with their corresponding singular vectors to obtain the approximated matrix $\mathrm{\widehat{W}_{\text{up}}}$ (Equation~\ref{eq:lr-apox}).
It is worth noting that the rank reduction raio candidate list includes $\Theta_0=0$, indicating that the weight matrix has the chance to remain unmodified.

With the searched rank reduction ratio, the weight matrix $\mathrm{W_{\text{up}}}$ in each CLIP layer is replaced and updated with its approximation. The \mname can be easily applied to different ViT backbones (Table~\ref{tab:backbone}), by replacing the model weight matrices with their low-rank approximations in a similar approach.
Then SeTAR detects the OOD data samples following MCM (Equation~\ref{eq:mcm_cal}), GL-MCM (Equation~\ref{eq:glmcm_cal}) or other scoring-based OOD detection method with the approximated model.
We provide an example procedure of the greedy search in Listing~\ref{lst:log} for better understanding.


\subsection{OOD Detection with \mnamec-enhanced Low-rank Adaptation}
\textbf{\mname} can be further combined with LoRA~\citep{hu2022lora} as a novel low-rank adaptation method for OOD detection, which we refer to as \textbf{\mnamec+FT}.
Specifically, we first apply \mname to the pre-trained CLIP model to obtain the reserved rank $r$ for each weight matrix $W$. Then we have
\vspace{-5pt}
\begin{align}
W&=\widehat{W}+B\times A\\
B&=\sum_{i=r+1}^{\min(m,n)} \sqrt{\sigma_i^{\downarrow}(W)} u_i\\
A&=\sum_{i=r+1}^{\min(m,n)} \sqrt{\sigma_i^{\downarrow}(W)} v_i^{\top}
\end{align}
where $\widehat{W}$ is the low-rank approximation of $W$ found by \mname, with $A$ and $B$ being the minor singular components. During finetuning, we keep $\widehat{W}$ frozen and only update the low-rank matrix $A$ and $B$. In this way, we retain the principle singular components in the original weight matrix and only update the minor singular components.Unlike LoRA, which evenly distributes the finetuning rank budget across all layers, \mnamec+FT adjusts the rank for each layer, resulting in more effective and efficient fine-tuning (Table~\ref{tab:compare_ft} and Figure~\ref{fig:loss_plot}). More details are provided in Section~\ref{sec:search_ft}.



\section{Experiments} \label{sec:exp}
\subsection{Experimental Setup} \label{sec:setup}

\paragraph{Dataset}
Following previous work \citep{ming2022delving,miyai2023zero}, we use two real-world datasets created from ImageNet1K~\citep{deng2009imagenet} and Pascal-VOC \citep{everingham2009pascal} as the ID datasets. For OOD datasets, we follow \citet{ming2022delving} to preprocess iNaturalist, SUN, Places and Texture, and follow \citet{miyai2023zero} to preprocess ImageNet22K and COCO data.
For finetune experiments, we follow \cite{miyai2023locoop} to use ImageNet1K as the ID dataset.
The detailed description and statistics of the datasets are provided in Appendix~\ref{appendix:data}.

\paragraph{Settings}
Following existing studies \citep{ming2022delving,miyai2023zero,miyai2023locoop}, we use CLIP ViT-B/16\footnote{\url{https://huggingface.co/openai/clip-vit-base-patch16}} \citep{radford2021learning} as our backbone. Both image and text encoders have 12 layers. More results with different backbones are in Section~\ref{sec:ablation}. The rank reduction ratio candidates range from 0 to 40\% in 5\% intervals. We use a temperature of $1$\footnote{Temperature is set to $1.0$ for the scaled CLIP logits, equivalent to the unscaled CLIP logits with a temperature of $100$. We adopt the unscaled setting in our implementation.}, unless stated otherwise. In all experiments, we use one CLIP text prompt: "a photo of a [CLASS],", where [CLASS] is the ID class name.
We set hyperparameters $\lambda$ (Equation~\ref{eq:locoop_loss}) and top-K (Equation~\ref{eq:ood_region}) according to the specific ID datasets and backbones. Detailed settings are in Table~\ref{tab:hyper_search}, with a sensitivity analysis in Section~\ref{sec:ablation}.
For \mnamec+FT and LoRA experiments, the learning rate and epoch number are set to $1e-2$ and $5$ for all experiments. The LoRA rank $r$ is set to match the trainable parameters of \mnamec+FT. Detailed settings are in Table~\ref{tab:hyper_ft}.
We report results from three runs with seeds 3, 4, 5\footnote{For \mname, the results are the same under different random seeds as it does not require training.}.
All experiments are conducted on a single NVIDIA RTX 4090 GPU. The time cost for low-rank approximation with CLIP-base on the ImageNet1K validation set is about 20 minutes.

\paragraph{Metrics}
We use the following metrics for evaluation. (1) the false positive rate (FPR95) for out-of-distribution (OOD) samples at a fixed true positive rate (TPR) of 95\% for in-distribution samples, with lower values targeting better performance; and (2) the area under the receiver operating characteristic curve (AUROC) for OOD samples, with higher values indicating better performance.

\paragraph{Baselines}
We evaluate \mname against MCM \citep{ming2022delving} and GL-MCM \citep{miyai2023zero}, state-of-the-art zero-shot OOD detection methods on CLIP. We also compare \mnamec+FT with fine-tuning baselines NPOS \citep{tao2023non}, CoOp \citep{zhou2021learning}, LoCoOp \citep{miyai2023locoop}, and LoRA \citep{hu2022lora}. More details are in Appendix~\ref{appendix:model}.

\subsection{Training-free Results}\label{sec:train-free}

\begin{table}[]
\vspace{-15pt}
\caption{\textbf{Training-free results of FPR95(FPR) and AUROC(AUC) compared to zero-shot baselines on CLIP-base.} \textbf{Bold} values represent the highest performance. $^\dagger$ is cited from \cite{miyai2023zero}, where $\diamond$ represents the absence of reporting in the paper. $^\ast$ denotes the result of our re-run. $-$ denotes the OOD dataset has overlapping categories with the ID dataset. We do not report standard deviations since no training is involved.}
\centering
\resizebox{\textwidth}{!}{
\begin{tabular}{lcccccccccccccccc}
\toprule
\multirow{2}{*}{Method} & \multicolumn{2}{c}{iNaturalist} & \multicolumn{2}{c}{SUN} & \multicolumn{2}{c}{Places} & \multicolumn{2}{c}{Texture} & \multicolumn{2}{c}{ImageNet22K} & \multicolumn{2}{c}{COCO}         & \multicolumn{2}{c}{\textbf{Average}} \\
\cmidrule(lr){2-3} \cmidrule(lr){4-5} \cmidrule(lr){6-7} \cmidrule(lr){8-9} \cmidrule(lr){10-11} \cmidrule(lr){12-13} \cmidrule(lr){14-15} \cmidrule(lr){16-17}
                        & FPR$\downarrow$   & AUC$\uparrow$  & FPR$\downarrow$       & AUC$\uparrow$       & FPR$\downarrow$ & AUC$\uparrow$        & FPR$\downarrow$ & AUC$\uparrow$ & FPR$\downarrow$       & AUC$\uparrow$       & FPR$\downarrow$  & AUC$\uparrow$ & FPR$\downarrow$      & AUC$\uparrow$    \\
\midrule
\multicolumn{17}{l}{\textbf{ImageNet1K}}                                                                                                                                                                                                                                                                     \\
\multicolumn{17}{l}{\textbf{MCM Score}}                                                                                                                                                                                                                                                                      \\
Vanilla MCM$^\dagger$    & 30.91          & 94.61          & 37.59          & 92.57          & 44.69          & 89.77          & 57.77          & 86.11          & -              & -              & -              & -              & 42.74             & 90.77            \\
Vanilla MCM$^\ast$      & 32.07          & 94.43          & 38.65          & 92.37          & 43.73          & 90.03          & 57.89          & 86.13          & -              & -              & -              & -              & 43.09             & 90.74            \\
\mname                   & \textbf{26.92} & \textbf{94.67} & \textbf{35.57} & \textbf{92.79} & \textbf{42.64} & \textbf{90.16} & \textbf{55.83} & \textbf{86.58} & -              & -              & -              & -              & \textbf{40.24}    & \textbf{91.05}   \\
\textbf{GL-MCM Score}   &                &                &                &                &                &                &                &                &                &                &                &                &                   &                  \\
Vanilla GL-MCM$^\dagger$ & 15.18          & 96.71          & 30.42          & 93.09          & 38.85          & 89.90          & 57.93          & 83.63          & -              & -              & -              & -              & 35.47             & 90.83            \\
Vanilla GL-MCM$^\ast$   & 15.34          & 96.62          & 30.65          & 93.01          & 37.76          & 90.07          & 57.41          & 83.73          & -              & -              & -              & -              & 35.29             & 90.86            \\
\mname                   & \textbf{13.36} & \textbf{96.92} & \textbf{28.17} & \textbf{93.36} & \textbf{36.80} & \textbf{90.40} & \textbf{54.17} & \textbf{84.59} & -              & -              & -              & -              & \textbf{33.12}    & \textbf{91.32}   \\
\midrule
\multicolumn{17}{l}{\textbf{Pascal-VOC}}                                                                                                                                                                                                                                                                     \\
\multicolumn{17}{l}{\textbf{MCM Score}}                                                                                                                                                                                                                                                                      \\
Vanilla MCM$^\dagger$    & 8.20           & 98.23          & 28.60          & 94.68          & $\diamond$     & $\diamond$   & 51.70          & 91.45          & 51.40          & 90.94          & 54.50          & 89.02          & 38.88             & 92.86            \\
Vanilla MCM$^\ast$      & 7.24           & 98.23          & 27.91          & 94.56          & 32.40          & 92.45          & 51.61          & 91.89          & 50.60          & 91.42          & 53.70          & 89.30          & 37.24             & 92.98            \\
\mname                    & \textbf{4.59}  & \textbf{98.71} & \textbf{24.91}         & \textbf{95.15}         & \textbf{28.46} & \textbf{93.21} & \textbf{40.44}           & \textbf{93.58}           & \textbf{48.25}             & \textbf{92.08}             & \textbf{48.10}          & \textbf{89.70}         & \textbf{32.46}                & \textbf{93.74}               \\
\multicolumn{17}{l}{\textbf{GL-MCM Score}}                                                                                                                                                                                                                                                                   \\
Vanilla GL-MCM$^\dagger$ & 4.20          & 98.71          & 23.10          & 94.66          & $\diamond$     & $\diamond$    & 43.00          & 92.84          & 41.00          & 92.38          & 44.30          & 90.48          & 31.12             & 93.81            \\
Vanilla GL-MCM$^\ast$   & 4.33           & 98.81          & 22.94          & 94.63          & 26.20          & 93.11          & 41.61          & 92.88          & 37.88          & 93.17          & 43.70          & 90.71          & 29.44             & 93.88            \\
\mname                    & \textbf{3.66}  & \textbf{98.96} & \textbf{21.93}         & \textbf{94.81}         & \textbf{25.04} & \textbf{93.62} & \textbf{20.35}           & \textbf{96.36}           & \textbf{31.47}             & \textbf{94.31}             & \textbf{40.70}          & \textbf{91.19}         & \textbf{23.86}                & \textbf{94.87} \\
\bottomrule
\end{tabular}
}
\label{tab:main}
\vspace{-15pt}
\end{table}

The training-free OOD detection performances are summarized in~Table~\ref{tab:main}.
Compared with zero-shot baselines, a salient observation is that on both MCM and GL-MCM, using \mname outperforms the vanilla method by a large margin across all OOD detection tasks.
For example, using Pascal-VOC as ID, \mname yields a \lyx{relatively} average reduction of 12.84\% FPR95 on MCM and 18.95\% FPR95 on GL-MCM. Considering that \mname is generally applicable and training-free, these results are very encouraging.
Comparing \mname with scoring function MCM and GL-MCM, \mnamec+GL-MCM performs better on all OOD detection tasks. However, the superiority of GL-MCM score over MCM appears to be contingent upon the choice of the model backbone. As evidenced in Table~\ref{tab:backbone}, \mnamec+MCM demonstrates superior performance with a \lyx{relatively} average FPR95 reduction of 8.30\% compared to \mnamec+GL-MCM with CLIP-large as the backbone on ImageNet1K.

\subsection{Fine-tuning Results}\label{sec:search_ft}

In this section, we compare \mnamec+FT with fine-tuning baselines, including NPOS \citep{tao2023non}, CoOp \citep{zhou2021learning}, LoCoOp \citep{miyai2023locoop} and LoRA \citep{hu2022lora}.
LoCoOp is the state-of-the-art prompt-learning OOD detection method on CLIP. LoRA is a representative parameter-efficient fine-tuning method.
Following previous work \citep{tao2023non, zhou2021learning, miyai2023locoop}, we report the results on the the ImageNet1K benchmark in Table~\ref{tab:compare_ft}. We observe that \mnamec+FT outperforms all baselines on both MCM and GL-MCM scoring functions.
For example, with CLIP-base as the backbone, \mnamec+FT achieves a \lyx{relatively} average FPR95 reduction of 3.97\% and 6.67\% compared to LoCoOp and LoRA.
Moreover, when scaled up to CLIP-large, \mnamec+FT outperforms LoCoOp and LoRA by 
\begin{wraptable}{r}{0.5\textwidth}
\vspace{-10pt}
\caption{\textbf{Fine-tuning results on ImageNet1K benchmark.} \textbf{Bold} values indicate the highest performance. $^\dagger$ is cited from \citet{tao2023non}. $^*$ denotes our re-run results, $\pm$ indicates the standard deviation from 3 runs.}
\centering
\resizebox{0.5\textwidth}{!}{
\begin{tabular}{lllllr}
\toprule
\multirow{2}{*}{\textbf{CLIP-base}} & \multicolumn{2}{c}{MCM Score}   & \multicolumn{2}{c}{GL-MCM Score} \\
                                        & \multicolumn{1}{c}{FPR95$\downarrow$}   & \multicolumn{1}{c}{AUROC$\uparrow$}  & \multicolumn{1}{c}{FPR95$\downarrow$}    & \multicolumn{1}{c}{AUROC$\uparrow$} \\
\midrule
NPOS$^\dagger$                         & 42.20             & 90.43             & 36.86              & 90.37              \\
CoOp$^\dagger$                         & 44.81             & 90.03             & 36.58              & 90.25              \\
LoCoOp$^\dagger$                       & 40.17             & 91.53             & 33.52              & 92.14              \\
LoCoOp$^*$                             & 39.76$_{\pm4.06}$ & 91.22$_{\pm0.52}$ & 34.14$_{\pm1.64}$  & 91.73$_{\pm0.17}$  \\
LoRA$^*$                               & 41.67$_{\pm0.14}$ & 90.85$_{\pm0.01}$ & 34.36$_{\pm0.11}$  & 90.88$_{\pm0.01}$  \\
SeTAR+FT                               & \textbf{38.77$_{\pm0.22}$}  & \textbf{91.55$_{\pm0.01}$}    & \textbf{32.19$_{\pm0.20}$}     & \textbf{92.31$_{\pm0.05}$}  \\
\midrule
\multirow{2}{*}{\textbf{CLIP-large}} & \multicolumn{2}{c}{MCM Score} & \multicolumn{2}{c}{GL-MCM Score} \\
                                        & \multicolumn{1}{c}{FPR95$\downarrow$}   & \multicolumn{1}{c}{AUROC$\uparrow$}   & \multicolumn{1}{c}{FPR95$\downarrow$}    & \multicolumn{1}{c}{AUROC$\uparrow$} \\
\midrule
LoCoOp$^*$                             & 40.74$_{\pm3.80}$ & 91.13$_{\pm0.79}$ & 46.74$_{\pm4.19}$  & 89.32$_{\pm0.80}$ \\
LoRA$^*$                               & 38.62$_{\pm0.07}$ & 91.66$_{\pm0.02}$ & 43.39$_{\pm0.01}$  & 89.76$_{\pm0.03}$ \\
SeTAR+FT                               & \textbf{34.75$_{\pm0.55}$}    & \textbf{92.86$_{\pm0.15}$}    & \textbf{37.05$_{\pm0.59}$}     & \textbf{91.83$_{\pm0.12}$} \\
\midrule
\multirow{2}{*}{\textbf{Swin-base}} & \multicolumn{2}{c}{MSP Score}   & \multicolumn{2}{c}{Energy Score} \\
                                        & \multicolumn{1}{c}{FPR95$\downarrow$}   & \multicolumn{1}{c}{AUROC$\uparrow$}   & \multicolumn{1}{c}{FPR95$\downarrow$}    & \multicolumn{1}{c}{AUROC$\uparrow$} \\
\midrule
LoRA$^*$                               & 57.02$_{\pm0.03}$ & 80.49$_{\pm0.01}$ & 62.17$_{\pm0.02}$  & 72.80$_{\pm0.00}$ \\
SeTAR+FT                               & \textbf{47.12$_{\pm0.42}$}    & \textbf{87.80$_{\pm0.44}$}    & \textbf{39.29$_{\pm0.57}$}     & \textbf{88.01$_{\pm0.51}$} \\
\bottomrule
\end{tabular}
}
\label{tab:compare_ft}
\vspace{-10pt}
\end{wraptable}
\lyx{relatively} 17.92\% and 12.45\% FPR95 on the same benchmark.
Similar results are observed on Swin Transformer~\citep{liu2021Swin}, where \mnamec+FT outperforms LoRA by \lyx{relatively} 17.36\% and 36.80\% FPR95 on MSP and Energy scoring functions, respectively. 
\lyx{The larger improvement on Swin Transformer may stem from its reliance on ImageNet training, making it prone to overfitting and weaker at OOD detection. Our method mitigates these issues, enhancing Swin's generalization to OOD instances.}
These results demonstrate the effectiveness and scalability of \mnamec+FT in improving the OOD detection performance.

Furthermore, \lyx{as shown in Figure~\ref{fig:loss_plot}, \mnamec+FT demonstrates faster convergence and lower loss than LoRA, especially in OOD loss, indicating that \mnamec+FT is more effective in adapting the pre-trained weights to the OOD detection task.}

\subsection{Ablation Study} \label{sec:ablation}
In this section, we conduct ablation studies with CLIP-base to understand our design choices.

\paragraph{Image v.s. Text modality}
Table~\ref{tab:modality} shows an ablation study on the modality involved in \mnamec. 
\begin{wraptable}{r}{0.5\textwidth}
\vspace{-10pt}
\caption{\textbf{Ablation study on modality.}}
\centering
\resizebox{0.5\textwidth}{!}{
\begin{tabular}{lcccccc}
\toprule
\multirow{2}{*}{Score} & \multicolumn{2}{c}{Vision} & \multicolumn{2}{c}{Text} & \multicolumn{2}{c}{Vision+Text} \\
\cmidrule(lr){2-3} \cmidrule(lr){4-5} \cmidrule(lr){6-7}
                       & FPR$\downarrow$ & AUC$\uparrow$ & FPR$\downarrow$ & AUC$\uparrow$ & FPR$\downarrow$ & AUC$\uparrow$ \\
\midrule
\multicolumn{7}{l}{\textbf{ImageNet1K}}                               \\
MCM                    & 40.27           & \textbf{91.24} & 42.78           & 90.50           & \textbf{40.24}  & 91.05          \\
GL-MCM                 & \textbf{32.97}           & \textbf{91.60} & 35.82           & 90.55           & 33.12  & 91.32          \\
\midrule
\multicolumn{7}{l}{\textbf{Pascal-VOC}}                               \\
MCM                    & 33.19           & 93.45          & 33.47           & 93.42           & \textbf{32.46}  & \textbf{93.74} \\
GL-MCM                 & 24.88           & 94.51          & 24.59           & 94.52           & \textbf{23.86}  & \textbf{94.87} \\
\bottomrule
\end{tabular}
}
\label{tab:modality}
\vspace{-10pt}
\end{wraptable}
As shown, the vision modality outperforms the text modality, 
\lyx{indicating the vision modality is more dominant in enhancing the model’s performance.}
When considering the vision modality alone and the combined vision+text modality, the latter either outperforms or achieves comparable average results to the former. Consequently, we make modifications to both the vision and text modalities in \mname to enhance overall performance.

\paragraph{Different Weight Types}

\begin{wraptable}{r}{0.5\textwidth}
\vspace{-11pt}
\caption{\textbf{Comparison results of \mname with and without considering projection matrix $\mathrm{W_{p}}$.}}
\centering
\resizebox{0.5\textwidth}{!}{
\begin{tabular}{lcccccc}
\toprule
\multirow{2}{*}{Score} & \multicolumn{2}{c}{Vanilla} & \multicolumn{2}{c}{\mname w $\mathrm{W_{p}}$} & \multicolumn{2}{c}{\mname w/o $\mathrm{W_{p}}$} \\
\cmidrule(lr){2-3} \cmidrule(lr){4-5} \cmidrule(lr){6-7}
                        & FPR$\downarrow$ & AUC$\uparrow$  & FPR$\downarrow$   & AUC$\uparrow$   & FPR$\downarrow$    & AUC$\uparrow$    \\
\midrule
\multicolumn{7}{l}{\textbf{ImageNet1K}}  \\
MCM                    & 43.09        & 90.74        & 41.79          & 90.74         & \textbf{40.24}  & \textbf{91.05} \\
GL-MCM                 & 35.29        & 90.86        & 34.30          & 91.24         & \textbf{33.12}  & \textbf{91.32} \\
\midrule
\multicolumn{7}{l}{\textbf{Pascal-VOC}}  \\
MCM                    & 37.24        & 92.98        & 35.94          & 93.32         & \textbf{32.46}  & \textbf{93.74} \\
GL-MCM                 & 29.44        & 93.88        & \textbf{23.34}          & 94.82         & 23.86  & \textbf{94.87} \\
\bottomrule
\end{tabular}
}
\label{tab:proj_w}
\vspace{-13pt}
\end{wraptable}
In this part, we present empirical evidence for modifying $\mathrm{W_{\text{up}}}$. We first compare the performance of \mname with different types of weight matrix in each Transformer layer, including $\mathrm{W_q}$, $\mathrm{W_k}$, $\mathrm{W_v}$, $\mathrm{W_o}$, $\mathrm{W_{\text{up}}}$ and $\mathrm{W_{\text{down}}}$. As shown in Figure~\ref{fig:weight_imagenet} and Figure~\ref{fig:weight_voc} of Appendx~\ref{appendix:exp}, the $X$-axis denotes the number of weight matrixes (layers) that we have searched, while the $Y$-axis is the average AUROC and FPR95. The results show that $\mathrm{W_{\text{up}}}$ consistently outperforms other weight matrices in terms of both AUROC and FPR95.
In addition to weight matrics in each transformer layer, CLIP has one projection matrix $\mathrm{W_{p}}$ on top of each encoder, which serves to project image/text representations into a shared space. 
In Table \ref{tab:proj_w}, we compare the performance of \mname with and without modifying $\mathrm{W_{p}}$. We search $\mathrm{W_p}$ first right before searching the image/text encoder. The results show that frozen $\mathrm{W_{p}}$ brings a \lyx{relatively} reduction of $4.20\%$ FPR95. Consequently, we keep $\mathrm{W_{p}}$ frozen in \mnamec.


\paragraph{Different Search Algorithms} \label{sec:dif-search}

At each step of the greedy search, \mname traverses the subsequent $\mathrm{W_{\text{up}}}$ in a predefined order and searches over different thresholds. We compare our method with two alternatives: modality-interleaved greedy search (MIS) and layer-exhaustive search (LES). MIS searches the image and text layers in an interleaved manner, while LES simultaneously searches over both layers and thresholds at each step. \mnamec-S, has linear complexity with respect to the number of model layers, similar to MIS, whereas LES has quadratic complexity.
\begin{wraptable}{r}{0.5\textwidth}
\vspace{-10pt}
\caption{\textbf{Results for different search algorithms.} Here LES, MIS and SeTAR-S stand for layer-exhaustive search, modality-interleave greedy search, and the search algorithm of \mnamec.}
\centering
\resizebox{0.5\textwidth}{!}{
\begin{tabular}{lcccccc}
\toprule
\multirow{2}{*}{Score} & \multicolumn{2}{c}{LES} & \multicolumn{2}{c}{MIS}         & \multicolumn{2}{c}{SeTAR-S}     \\
\cmidrule(lr){2-3} \cmidrule(lr){4-5} \cmidrule(lr){6-7}
                        & FPR$\downarrow$ & AUC$\uparrow$ & FPR$\downarrow$     & AUC$\uparrow$     & FPR$\downarrow$     & AUC$\uparrow$     \\
\midrule
\multicolumn{7}{l}{\textbf{ImageNet1K}}  \\
MCM                    & 41.99      & 90.78      & 40.55          & 91.00          & \textbf{40.24} & \textbf{91.05} \\
GL-MCM                 & 33.90      & 91.08      & 33.36          & 91.29          & \textbf{33.12} & \textbf{91.32} \\
\midrule
\multicolumn{7}{l}{\textbf{Pascal-VOC}}  \\
MCM    & 35.11  & 93.60  & 33.93          & 93.58          & \textbf{32.46} & \textbf{93.74} \\
GL-MCM & 24.48  & 94.57  & \textbf{22.87} & 94.84          & 23.86          & \textbf{94.87} \\
\bottomrule
\end{tabular}
}
\label{tab:dif-search}
\vspace{-15pt}
\end{wraptable}
Table~\ref{tab:dif-search} presents the comparison results. \mnamec-S demonstrates better overall performance than MIS. Notably, MIS encounters limitations when the image and text towers have different layer counts (e.g., CLIP-large with 24 image layers and 12 text layers). Therefore, we choose \mnamec-S for better generalization.
Compared to LES, \mnamec-S performs better in terms of both FPR95 and AUROC, as LES's locally optimal algorithm may not achieve a global optimal solution. These results validate the superiority of our top-to-bottom layer search strategy.


\paragraph{Different Prune Strategies}
Inspired from SVD, \mname modify the model weights by pruning the minor singular components, and retains the principle components that contribute the most to the model's performance. To validate this design, we compare \mname with two alternatives: principal component pruning and random pruning pruning. Principal component takes 
\begin{wraptable}{r}{0.5\textwidth}
\vspace{-10pt}
\caption{\textbf{Results for different pruning strategies.}}
\centering
\resizebox{0.5\textwidth}{!}{
\begin{tabular}{llcccccc}
\toprule
\multirow{2}{*}{Score} & \multicolumn{2}{c}{Principle} & \multicolumn{2}{c}{Random} & \multicolumn{2}{c}{Minor} \\
\cmidrule(lr){2-3} \cmidrule(lr){4-5} \cmidrule(lr){6-7}
                       & FPR$\downarrow$ & AUC$\uparrow$ & FPR$\downarrow$ & AUC$\uparrow$ & FPR$\downarrow$ & AUC$\uparrow$ \\
\midrule
\multicolumn{7}{l}{\textbf{ImageNet1K}}  \\
MCM                    & 43.09    & 90.74 & 43.09   & 90.74  & \textbf{40.24} & \textbf{91.05}  \\
GL-MCM                 & 35.29    & 90.86 & 35.29   & 90.86  & \textbf{33.12} & \textbf{91.32}  \\
\midrule
\multicolumn{7}{l}{\textbf{Pascal-VOC}}  \\
MCM                    & 38.20    & 92.44 & 33.57   & 93.09  & \textbf{32.46} & \textbf{93.74} \\
GL-MCM                 & 25.36    & 93.67 & 26.20   & 94.66  & \textbf{23.86} & \textbf{94.87} \\
\bottomrule
\end{tabular}
}
\label{tab:prune_type}
\vspace{-15pt}
\end{wraptable}
the opposite approach, retaining minor components and pruning major ones. Random pruning, on the other hand, prunes weights randomly.
As shown in Table~\ref{tab:prune_type}, principle pruning suffers from a significant performance drop compared to \mname, while random pruning performs slightly better than principle pruning. These results demonstrate the effectiveness of \mname's design choice in pruning the minor components.

\paragraph{Sensitivity Analysis on $\lambda$ and top-K}

In this section, we present the sensitivity analysis of the hyperparameters $\lambda$ (Figure~\ref{fig:appendix_lambda_abs}) and top-K (Figure~\ref{fig:appendix_topk_abs}).
As observed in Figure~\ref{fig:appendix_lambda_abs}, the average AUROC remains stable at lower values and slightly decreases as $\lambda$ increases for both SeTAR+MCM and SeTAR+GL-MCM. Notably, the optimal setting of $\lambda$ may vary depending on the model backbone, with our experiments indicating that CLIP-large may require a larger $\lambda$ than CLIP-base.
\lyx{Despite this variation, the $\lambda$ parameter demonstrates strong transferability across datasets for the same backbone. Swapping the optimal $\lambda$ between ImageNet1K and Pascal-VOC has a minimal performance impact, consistently outperforming the vanilla method. With the VOC-optimized $\lambda$ on ImageNet1K, CLIP-base achieves an FPR95 of 40.91 and AUROC of 91.02, and CLIP-large reaches 46.73 FPR95 and 91.81 AUROC. Conversely, using the ImageNet1K-optimized $\lambda$ on Pascal-VOC, CLIP-base achieves 33.18 FPR95 and 93.65 AUROC, while CLIP-large attains 44.39 FPR95 and 92.3 AUROC.}

\lyx{Top-K controls the number of OOD regions considered in LoCoOp loss: higher values include more OOD regions, with top-K equal to the number of ID classes covering all OOD regions, and top-K set to 0 focusing solely on ID loss. 
The optimal top-K depends on the number of ID categories, making it non-transferable across datasets. However, \mname remains robust to top-K variations, as shown in Figure~\ref{fig:appendix_topk_abs}, except at extreme values (0 or the maximum number of classes).
We recommend setting top-K to around 30\% of the total categories, such as 300 for ImageNet1K and 4 for Pascal-VOC. For the Swin-base model, top-K at 300 on ImageNet1K yields an FPR95 of 56.82 and AUROC of 85.68 with MSP, and an FPR95 of 52.56 and AUROC of 84.51 with Energy.}


\subsection{Analyses} \label{sec:analyses}


\paragraph{Can \mname Improve Image Classification?}

\begin{wraptable}{r}{0.5\textwidth}
\vspace{-13pt}
\caption{\textbf{Image classification results with different methods.} We use ImageNet1K (IN1K) as ID dataset. $^*$ denotes the results of our re-run. The results are averaged over 3 runs.}
\centering
\resizebox{0.5\textwidth}{!}{
\begin{tabular}{lccccc}
\toprule
Method             & IN1K     & SUN            & Places         & Texture        & \textbf{Average} \\
\midrule
Vanilla CLIP$^*$ & 64.07          & 75.77          & 45.65          & 43.60          & 57.27            \\
LoCoOp$^*$   & 64.93          & 75.89          & 46.47          & 37.79          & 56.27            \\
LoRA$^*$     & 65.43          & 76.86          & 46.58          & \textbf{43.98}          & 58.21            \\
SeTAR        & 63.97          & 75.50          & 45.81          & 43.76 & 57.26            \\
SeTAR+FT     & \textbf{67.02} & \textbf{77.94} & \textbf{46.64} & 43.28          & \textbf{58.72}   \\
\bottomrule
\end{tabular}
}
\label{tab:classfiy_acc}
\vspace{-10pt}
\end{wraptable}

To evaluate the impact of \mname and \mnamec+FT on classification accuracy, we present our results on ID dataset ImageNet1K and OOD datasets SUN, Places and Texture in Table~\ref{tab:classfiy_acc}\footnote{We do not report classification accuracy on iNaturalist as we failed to match the labels for the OOD test set.}.
\mname effectively maintains the average accuracy, with minor variations observed across different datasets. Among the fine-tuned baselines, LoCoOp exhibits a 1\% decrease in accuracy compared to Vanilla CLIP, whereas LoRA shows an improvement of 0.94\%.
Notably, \mnamec+FT surpasses both baselines, improving the average accuracy by 1.45\% compared to Vanilla CLIP. These results highlight the efficacy of \mname and \mnamec+FT in improving OOD detection without compromising classification accuracy.

\paragraph{\mname is Effective on Different \lyx{Architectures and Score Functions}}

\lyx{We expand on Table~\ref{tab:main} with results on ViT and CNN backbones and various score functions. For ViT-based models, we evaluate OOD detection using CLIP-large\footnote{\url{https://huggingface.co/openai/clip-vit-large-patch14}} and Swin Transformer\footnote{\url{https://huggingface.co/microsoft/swinv2-base-patch4-window16-256}}~\citep{liu2021Swin}, alongside CLIP-base. The Swin Transformer~\citep{liu2022swin} is trained on ImageNet1K. Since it lacks a text encoder, we apply \mname to the image ViT only. For Swin Transformer, we use two common scoring functions: MSP~\citep{hendrycks2016baseline}, which leverages softmax confidence, and the Energy score~\citep{liu2020energy}, with $T=0.1$ for OOD detection. We also integrate CLIP-base }
\begin{wraptable}{r}{0.5\textwidth}
\vspace{-10pt}
\caption{\textbf{Results for different \lyx{ViT} backbones.}}
\centering
\resizebox{0.5\textwidth}{!}{
\begin{tabular}{llcccc}
\toprule
\multirow{2}{*}{Backbone} & \multirow{2}{*}{Score} & \multicolumn{2}{c}{Vanilla Method} & \multicolumn{2}{c}{\mnamec} \\
\cmidrule(lr){3-4} \cmidrule(lr){5-6}
                                     &                        & FPR$\downarrow$ & AUC$\uparrow$ & FPR$\downarrow$     & AUC$\uparrow$  \\
\midrule
\multicolumn{6}{l}{\textbf{ImageNet1K}}                               \\
CLIP-base  & \lyx{NegLabel} & 25.40 & 94.21 & \textbf{23.09} & \textbf{94.48
}   \\
CLIP-large & MCM    & 37.19 & 91.73 & \textbf{36.26} & \textbf{91.92} \\
CLIP-large & GL-MCM & 40.65 & 89.98 & \textbf{39.54} & \textbf{90.22} \\
Swin-base  & MSP    & 59.25 & 84.12 & \textbf{56.05} & \textbf{85.77} \\
Swin-base  & Energy & 65.01 & 76.10 & \textbf{51.61} & \textbf{84.42} \\
\midrule
\multicolumn{6}{l}{\textbf{Pascal-VOC}}                               \\
CLIP-large & MCM    & 52.21 & 91.68 & \textbf{42.57} & \textbf{92.91} \\
CLIP-large & GL-MCM & 43.96 & 92.45 & \textbf{31.12} & \textbf{94.00} \\
\bottomrule
\end{tabular}
}
\label{tab:backbone}
\vspace{-15pt}
\end{wraptable}
\lyx{with the NegLabel score function~\citep{jiang2024negative}, which uses large-scale negative labels.
As shown in Table~\ref{tab:backbone}, \mname consistently outperforms baselines across all backbones and scoring functions, significantly reducing FPR95 by relatively 20.61\% with the Energy score on Swin Transformer. These results demonstrate \mname's effectiveness in improving OOD detection for unimodal image encoders, with further confirmation from \mnamec+FT results (Table~\ref{tab:compare_ft}) across different model backbones.}

\begin{wraptable}{r}{0.5\textwidth}
\vspace{-10pt}
\caption{\lyx{\textbf{Results on ResNet50}. We use ImageNet1K as the ID dataset. $^\dagger$ is cited from \cite{djurisic2023extremely}.}}
\centering
\resizebox{0.5\textwidth}{!}{
\begin{tabular}{lll|lll}
\toprule
Method   & FPR↓  & AUC↑  & Method & FPR↓  & AUC↑  \\
\midrule
Softmax$^\dagger$ & 66.95 & 81.99 & ASH-P$^\dagger$ & 50.32 & 89.04 \\
Energy$^\dagger$  & 58.41 & 86.17 & ASH-B$^\dagger$ & 22.73 & 95.06 \\
ReAct$^\dagger$   & 31.43 & 92.95 & ASH-S$^\dagger$ & 22.80 & 95.12 \\
DICE$^\dagger$    & 34.75 & 90.77 & SeTAR  & \textbf{22.38} & \textbf{95.25} \\
\bottomrule
\end{tabular}
}
\label{tab:resnet50}
\vspace{-15pt}
\end{wraptable}

\lyx{We further explore SeTAR's potential on CNN architecture, and compare it with methods such as Softmax, Energy~\citep{Wu2023EnergybasedOD}, ReAct~\citep{sun2021react}, DICE~\citep{10.1007/978-3-031-20053-3_40}, and ASH~\citep{djurisic2023extremely} on ResNet50\footnote{\url{https://download.pytorch.org/models/resnet50-19c8e357.pth}}.
Since ResNet lacks local features for OOD loss, we conduct experiments using only ID loss. We apply low-rank approximation to the in- and out-feature dimensions of the convolutional layers, combined with ASH for search. As shown in Table~\ref{tab:resnet50}, \mname establishes new state-of-the-art results on ResNet, demonstrating its effectiveness across both ViT and CNN architectures.
}

\paragraph{\lyx{Near-OOD Results}} 

\begin{wraptable}{r}{0.5\textwidth}
\vspace{-13pt}
\caption{\lyx{\textbf{Near-OOD results on CLIP-base}.}}
\centering
\resizebox{0.5\textwidth}{!}{
\begin{tabular}{llcccc}
\toprule
\multirow{2}{*}{Method} & \multirow{2}{*}{Category} & \multicolumn{2}{c}{MCM Score} & \multicolumn{2}{c}{GL-MCM Score} \\
\cmidrule(lr){3-4} \cmidrule(lr){5-6}
                                     &                        & FPR$\downarrow$ & AUC$\uparrow$ & FPR$\downarrow$     & AUC$\uparrow$  \\
\midrule
Vanilla & Training-Free    & 89.28 & 63.88  & 85.62 & 67.63 \\
\mname  & Training-Free    & 88.29 & 64.20 & \textbf{84.03}  & 68.29 \\
LoCoOp  & Training-Free    & 89.72 & 63.45  & 86.79  & 65.93  \\
LoRA    & Finetuning       & 88.52 & 65.38  & 84.39  & 68.85  \\
SeTAR+FT  & Finetuning     & \textbf{87.16} & \textbf{68.13} & 84.72 & \textbf{70.42} \\
\bottomrule
\end{tabular}
}
\label{tab:near_ood}
\vspace{-20pt}
\end{wraptable}

\lyx{To further evaluate SeTAR's performance on diverse OOD tasks, we test it on a more challenging near-OOD setting using ImageNet1K as the ID dataset and SSB-Hard \citep{vaze2022openset} as the OOD dataset. As shown in Table~\ref{tab:near_ood}, SeTAR and SeTAR+FT outperform the baselines, demonstrating superior performance in near-OOD scenarios.}

\section{Related Work} \label{sec:related_work}

\paragraph{Out-of-Distribution Detection}

Previous work explores OOD detection with unimodal \citep{devries2018learning,hendrycks2016baseline,hu2021uncertainty,zheng2020ooddialog,zhou-etal-2021-contrastive} and multimodal \citep{fort2021exploring,ming2022delving,tao2023non,miyai2023locoop} models. Numerous methodologies \citep{lee2018simple,huang2021importance,sun2022out,haoqi2022vim,Wu2023EnergybasedOD} have been developed to tackle OOD detection in computer vision.
Existing CLIP-based OOD detection methods include zero-shot \citep{fort2021exploring,ming2022delving,miyai2023zero,Dai2023ExploringLL,Wang2023CLIPNFZ,jiang2024negative} and fine-tuning \citep{Ming_2023,tao2023non,miyai2023locoop}. Zero-shot methods like MCM \citep{ming2022delving} and GL-MCM \citep{miyai2023zero} don't require in-distribution training data but may perform suboptimally due to domain gaps. 
\lyx{Other approaches integrate external knowledge. For example, CLIPN \citep{Wang2023CLIPNFZ} pre-trains a novel NO-encoder on the CC-3M dataset \citep{sharma2018conceptual} to empower CLIP's "no" logic for zero-shot evaluation. NegLabel \citep{jiang2024negative} demonstrates better performance than CLIPN by introducing large-scale negative labels for enhanced label scoring.}
Fine-tuning methods \citep{Ming_2023,tao2023non,miyai2023locoop} improve OOD detection by adapting to in-distribution data but risk damaging the pretraining representations, needing careful training strategies.
\lyx{
CNN-based OOD detection methods, including ReAct \citep{sun2021react}, ASH \citep{djurisic2023extremely}, DICE \citep{10.1007/978-3-031-20053-3_40}, CIDER \citep{ming2023how}, PALM \citep{PALM2024}, and Hopfield Boosting \citep{hofmann2024energybasedhopfieldboostingoutofdistribution}, have also demonstrated strong results. However, methods like ReAct and ASH rely on the assumption that ID and OOD images produce distinct activations in models trained on ID data. This assumption does not hold in large-scale pre-trained models like CLIP, where activations for ID and OOD images are not significantly different, limiting the effectiveness of such approaches in enhancing CLIP’s zero-shot OOD detection capabilities.
SeTAR, in contrast, offers high compatibility with various scoring functions (e.g., MCM, GL-MCM, MSP, Energy), multiple model backbones (e.g., CLIP, Swin, ResNet), and advanced OOD techniques such as NegLabel. Designed to be both lightweight and efficient, SeTAR addresses the demand for resource-efficient solutions in OOD detection.
}

\paragraph{Low-rank Approximations of Weight Matrices}
Neural networks trained with over-parameterization often exhibit low-rank properties \citep{oymak2019generalization}. These properties are utilized in both model training \citep{Povey2018SemiOrthogonalLM,hu2022lora} and post-hoc processing \citep{Hajimolahoseini2021CompressingPL,sharma2023truth}. In training, some works \citep{sainath2013low,Zhang2014ExtractingDN,Zhao2016LowrankPD} impose low-rank constraints, while LoRA \citep{hu2022lora} adapts pretrained LLMs to downstream tasks using trainable low-rank matrices.
For post-hoc processing, pruning methods \citep{Yu2017OnCD,Hajimolahoseini2021CompressingPL} reduce weight matrix ranks by retaining top-K components from SVD. While pruning preserves model behavior, performance declines with increased intervention. 
\lyx{LASER \citep{sharma2023truth} focuses on pruning individual layers to enhance factual answering capabilities. It utilizes a simple greedy search strategy on a validation set, which is not applicable for OOD detection due to the absence of a validation set. In contrast, our approach introduces a selective rank reduction strategy specifically tailored for OOD detection. We systematically analyze and compare different greedy search techniques, evaluating their effectiveness across various layers and model backbones.}



\section{Conclusion} \label{sec:conclusion}
We propose \mname, a simple and effective OOD detection method using post-hoc low-rank approximation on weight matrices $\mathrm{W_{\text{up}}}$ with a top-down, image-to-text greedy search. \mname offers several advantages: (1) training-free, (2) scalable to unimodal and multimodal models, and (3) complementary to existing OOD scoring functions.
Building on \mname, we introduce \mnamec-FT, a finetuning method that adapts the model to in-distribution data for improved OOD detection. We evaluate \mname and \mnamec-FT on large-scale benchmarks, including ImageNet1K and Pascal-VOC. Results show that both achieve state-of-the-art OOD detection performance. We hope our work inspires further research and contributes to more robust and reliable models.

\section*{Acknowledgements}

This project was supported by National Natural Science Foundation of China (No. 62306132, No. 62106138). We thank the anonymous reviewers for their insightful feedbacks on this work.




\bibliography{custom}
\bibliographystyle{neurips_2024}

\clearpage
\newpage

\appendix
\onecolumn

\section{Impact Statements}\label{sec:impact_state}
\paragraph{Limitation}
While we demonstrate the effectiveness of our method on OOD detection, we acknowledge that our work has several limitations. First, despite we show the robustness of our method to hyperparameters, the optimal hyperparameters may vary across different model backbones. Future work is needed to explore the autonomous selection of hyperparameters.
Second, we design \mnamec+FT in a simple and straightforward manner, which may not be the most efficient or effective way to adapt the model to the ID downstream data. More sophisticated strategies for model adaptation are worth exploring in future research.
Third, we only conduct experiments to detect visual OOD inputs and ignore inputs in other modalities such as textual, audio and video. This is primarily because our model is based on CLIP. Exploring the development of OOD detectors across diverse modalities remains an active research topic for future investigation.

\paragraph{Ethical Considerations}
Our study addresses the challenge of OOD detection through low-rank approximation, which is particularly relevant for ensuring the reliability and trustworthiness of vision-and-language pre-trained models. Future investigations on fairness, privacy and transparency neural-based models should be encouraged to mitigate the existing data biases and safety problems for a responsible, helpful and trustworthy AI system in diverse real-world applications.


\paragraph{Future Societal Consequences}
Our proposed \mname achieves impressive OOD detection performance, which is beneficial to various real-world machine learning applications, such as healthcare and autonomous vehicles.
The identification of anomalies or unexpected data points is crucial for decision-making and risk management with AI models.
A better OOD detector facilitates the development of trustworthy machine-learning models that can reject unknown data inputs and help alleviate the hallucination problem. Moreover, better OOD detectors like \mname can help to select and label the unfamiliar data samples to further train a stronger model in the wild. 

\section{Loss Function} \label{sec:locoop_loss}
To improve the model's OOD detection ability, it is crucial to define a loss function that pushes OOD samples far from ID samples while keeping ID samples close to each other. However, since OOD samples are unavailable during development, we address this issue by using the LoCoOp loss~\citep{miyai2023locoop} for both \mname and \mnamec+FT. The main idea is to create pseudo OOD features with ID-irrelevant nuisances (e.g., backgrounds) in CLIP’s local features.

Specifically, we divide the image into patches, represented by the set of all patch indices $I = \{0,1,2,\ldots, H \times W - 1\}$, where $H$ and $W$ denote the height and width of the patch features.
Next, we compute the cosine similarity between the image patch features $p_i^v$ and the text features $h^t_c$ of the image label. The classification prediction probabilities for each patch $i$ are then given by:
\begin{equation}
    p_i(y=m | \mathbf{x}) = \frac{\exp(\mathrm{cos\_sim}(p_i^v, h^t_c) / \tau^\prime)}{\sum_{c=1}^K \exp(\mathrm{cos\_sim}(p_i^v, h^t_c) / \tau^\prime)}
\end{equation}

For a given image patch related to an ID category, the corresponding ID label should be among its top-K predictions. Conversely, for patches unrelated to the ID label, such as background regions, the ID label should be excluded from the top-K predictions. Based on this intuition, the indices of ID-irrelevant regions within an image are defined by Equation~\ref{eq:ood_region}, where $\text{rank}(p_i(y = \mathbf{y} | \mathbf{x}))$ denotes the rank of the true class $\mathbf{y}$ among all ID classes, and K is the hyperparameter.
\begin{equation}
    J = \{ i \ | \ \text{rank}(p_i(y = \mathbf{y} | \mathbf{x})) > \text{K} \}
\label{eq:ood_region}
\end{equation}

After identifying out-of-distribution (OOD) regions, it is expected that their image features will differ significantly from the ID text embeddings. To enhance this distinction, entropy maximization is employed to increase the entropy of $p_j(y|\mathbf{x})$, where $p_j$ denotes the classification prediction probabilities for region $j \in J$. The entropy maximization is formally defined as follows:
\begin{equation}
\mathcal{L}_{\text{ood}} = -H(p_j)
\end{equation}

Here, $H(\cdot)$ represents the entropy function. The overall loss function combines the ID loss (cross-entropy loss for ID predictions) with the OOD loss. Here $\lambda$ is the hyperparameter that regulates the proportion of the OOD loss.
\begin{equation}
\label{eq:locoop_loss}
\mathcal{L} = \mathcal{L}_{\text{id}} + \lambda \mathcal{L}_{\text{ood}}
\end{equation}

\section{Data}\label{appendix:data}

\begin{table}[H]
    \centering
    \vspace{-15pt}
    \caption{\textbf{The statistics of the dataset used in this paper.} `ID' and `OOD' denote in-distribution and out-of-distribution, respectively.}
    \label{tab:app-dataset}
    \resizebox{0.65\columnwidth}{!}{
    \begin{tabular}{cccc}
    \toprule
    Data         & Type & Valid Size & Test Size  \\ \midrule
    ImageNet1K \citep{deng2009imagenet}  & ID   &  1,000  &  50,000  \\
    Pascal-VOC \citep{everingham2009pascal}  & ID   &  94  &  906  \\
    iNaturalist \citep{van2018inaturalist}  & OOD   &  0  &  10,000  \\
    SUN \citep{xiao2010sun}         & OOD   &  0  &  10,000  \\
    Places \citep{zhou2017places}      & OOD   &  0  &  10,000  \\
    Texture \citep{cimpoi2014describing}     & OOD   &  0  &  5,640  \\
    ImageNet22K \citep{russakovsky2015imagenet}     & OOD   &  0  & 18,335  \\
    COCO  \citep{lin2014microsoft}    & OOD   &  0  & 1,000  \\
    \bottomrule
\end{tabular}
}
\vspace{-5pt}
\end{table}

We use two real-world datasets created from ImageNet1K~\citep{deng2009imagenet} and Pascal-VOC \citep{everingham2009pascal} as the ID dataset. We use ImageNet-1K validation set as the ID test set following \citet{ming2022delving}, and preprocess Pascal-VOC following \citet{miyai2023zero}.
we build two ID validation sets for low-rank approximation.
The ID validation set of ImageNet1K is collected by sampling one image for each label from the ImageNet1K training set.
For Pascal-VOC, For Pascal-VOC, We randomly sample 10\% images as the ID validation set and leave the rest as the ID test set.

For OOD datasets, we follow \citet{ming2022delving} to preprocess iNaturalist, SUN, Places and Texture, and follow \citet{miyai2023zero} to preprocess ImageNet22K and COCO data. We only evaluate the OOD datasets that have no overlapping categories as the ID dataset.

We provide more details about the datasets used in our experiments, in terms of data sources, preprocessing, and the statistics for each dataset, as shown in Table~\ref{tab:app-dataset} and below.

\paragraph{ImageNet1K} We use the ImageNet-1000 (ILSVRC2012) \citep{deng2009imagenet} dataset for ID validation and testing. The original dataset contains 1.2 million training images and 50,000 validation images from 1000 classes, and is widely used for image classification. We follow \citet{ming2022delving} to construct the ImageNet1K ID test set from the validation set. Additionally, we curate an ImageNet1K ID validation set from the training set by randomly selecting one image for each label.


\paragraph{Pascal-VOC} The Pascal VOC (Visual Object Classes) \citep{everingham2009pascal} dataset is a benchmark dataset widely used in computer vision, featuring annotated images across multiple object categories. We use the Pascal-VOC subset collected by \citet{miyai2023zero} as the ID dataset, each image has single-class ID objects and one or more OOD objects. The ID validation and test set are split by 1:9 for each class, resulting in 94 and 906 images, respectively.

\paragraph{iNaturalist} iNaturalist \citep{van2018inaturalist} is a biodiversity dataset containing millions of labeled images of plants, animals, and insects. \citet{ming2022delving} construct a subset with 10,000 images by de-duplicating concepts overlapped with ID datasets.

\paragraph{Places} Places \citep{zhou2017places} is a scene-centric database with 205 scene categories and 2.5 million images. We use the SUN subset collected by \citet{ming2022delving} as the OOD test set, which contains 10,000 images that are not overlapped with the ID classes.

\paragraph{SUN} SUN (Scene UNderstanding) \citep{xiao2010sun} is a comprehensive collection of labeled images representing a diverse range of indoor and outdoor scenes. We use the SUN subset collected by \citet{ming2022delving} as the OOD test set, which contains 10,000 images that are not overlapped with the ID classes.

\paragraph{Texture} The Texture dataset (DTD) \citep{cimpoi2014describing} comprises 5640 images categorized into 47 terms inspired by human perception, aimed at replicating human-like texture recognition in machines. Again, we use the subset collected by \citet{ming2022delving} as the OOD test set.

\paragraph{ImageNet22K} The ImageNet-22K dataset \citep{russakovsky2015imagenet}, formerly known as ImageNet-21K, addresses the underestimation of its additional value compared to the standard ImageNet-1K pretraining, aiming to provide high-quality pretraining for a broader range of models. We use the filtered subset collected by \citet{wang2021can} as the OOD test set for MC-COCO and Pascal-VOC ID test sets.

\paragraph{COCO} \citet{miyai2023zero} curated an MS-COCO OOD test set (COCO for short) with 1,000 images that are not overlapped with the Pascal-VOC ID classes, which we use as OOD testing data for Pascal-VOC ID test set.

\section{Fine-tune Baselines}\label{appendix:model}
We compare \mnamec+FT with 4 finetuning-based baselines. These baselines include:
\begin{itemize}
\item \textbf{NPOS.} NPOS \citep{tao2023non} generates virtual anomalies in low-probability regions of ID data without relying on distribution assumptions, enhancing discrimination during training.
\item \textbf{CoOp.} CoOp \citep{zhou2021learning} optimizes prompts for vision-language models with learnable context vectors for efficient few-shot learning.
\item \textbf{LoCoOp.} LoCoOp \citep{miyai2023locoop} improves upon CoOp by leveraging CLIP's local features to better distinguish between ID and OOD samples, achieving higher detection accuracy with less training data. We follow the official code\footnote{\url{https://github.com/AtsuMiyai/LoCoOp}} to prepare and fine-tune the LoCoOp with CLIP-base and CLIP-large. Follow \citet{miyai2023locoop}, the top-K, $\lambda$, learning rate and epoch num are set to 200, 0.25, 0.002 and 50. Temperature is set to 1 and the text prompt is initiated with ``\texttt{X X X X X X X X X X X X X X X X [CLASS]}'', where [CLASS] is the ID class name. We average the results from 3 seeds finetuned with 1-shot ImageNet1K valid data.

\item \textbf{LoRA.} LoRA~\citep{hu2022lora} is a low-rank adaptation method that injects trainable low-rank decomposition matrices into the pre-trained model to adapt to downstream tasks.
We apply low-rank adaptation to the same weight type as \mnamec+FT, the rank of each layer is set to match the trainable parameters of \mnamec. Details settings can be found in Table~\ref{tab:hyper_ft}.

\end{itemize}


\section{HyperParameters Settings}\label{sec:hyper_settings}

The hyperparameters for \mname are shown in Table~\ref{tab:hyper_search}. And the hyperparameters for \mnamec+FT and LoRA are shown in Table~\ref{tab:hyper_ft}.

\begin{table}[htbp]
\caption{\textbf{Hyperparameters for \mname.} Temperature is set to 1 except for Swin-base with Energy score, where it is set to 0.1.}
\centering
\resizebox{0.45\textwidth}{!}{
\begin{tabular}{llcc}
\toprule
Backbone                    & Dataset    & $\lambda$ & top-K \\
\midrule
\multirow{2}{*}{CLIP-base}  & ImageNet1K & 0.10   & 300   \\
                            & Pascal-VOC & 0.05   & 4     \\
\midrule
\multirow{2}{*}{CLIP-large} & ImageNet1K & 0.50   & 300   \\
                            & Pascal-VOC & 0.30   & 6     \\
\midrule
Swin-base                   & ImageNet1K & 0.01   & 700   \\
\bottomrule
\end{tabular}
}
\label{tab:hyper_search}
\end{table}

\begin{table}[htbphtbp]
\caption{\textbf{Hyperparameters for \mnamec+FT and LoRA on ImageNet1K.} Temperature is set to 1 except for Swin-base with Energy score, which is set to 0.1.}
\centering
\resizebox{0.8\textwidth}{!}{
\begin{tabular}{lcccccc}
\toprule
Backbone   & $\lambda$ & top-K & LR   & Epoch  & Rank for LoRA & Alpha for LoRA \\
\midrule
CLIP-base  & 0.10   & 300   & 0.01 & 5  & 32            & 16             \\
CLIP-large & 0.50   & 300   & 0.01 & 5  & 64            & 16             \\
Swin-base  & 0.01   & 700   & 0.01 & 5  & 112           & 16             \\
\bottomrule
\end{tabular}
}
\label{tab:hyper_ft}
\end{table}

\section{More Detailed Experiment Results}\label{appendix:exp}

In this section, we present additional detailed results from the main paper. This includes the detailed results of fine-tuned baselines on the ImageNet1K benchmark in Table~\ref{tab:finetune_appendix}; detailed ablation results on modality, $\mathrm{W_{p}}$, $\lambda$, and top-K in Table~\ref{tab:modality_appendix}, Table~\ref{tab:proj_w_appendix}, Table~\ref{tab:lambda_appendix}, and Table~\ref{tab:topk_appendix}; and detailed results of \mname with different search algorithms, prune strategies and backbones in Table~\ref{tab:dif-search_appendix}, Table~\ref{tab:prune_type_appendix}, Table~\ref{tab:backbone_appendix} and Table~\ref{tab:resnet50_appendix}.

\begin{table}[htbp]
    \caption{\textbf{Detail results of FPR95(FPR) and AUROC(AUC) compared with fine-tuned baselines on ImageNet1K benchmark.} $^\dagger$ is cited from \citet{tao2023non}. $^*$ denotes the results of our re-run.}
\centering
\resizebox{\textwidth}{!}{
\begin{tabular}{lcccccccccc}
\toprule
\multirow{2}{*}{Method} & \multicolumn{2}{c}{iNaturalist} & \multicolumn{2}{c}{SUN}         & \multicolumn{2}{c}{Places}      & \multicolumn{2}{c}{Texture}     & \multicolumn{2}{c}{\textbf{Average}} \\
\cmidrule(lr){2-3} \cmidrule(lr){4-5} \cmidrule(lr){6-7} \cmidrule(lr){8-9} \cmidrule(lr){10-11}
                        & FPR$\downarrow$ & AUC$\uparrow$  & FPR$\downarrow$ & AUC$\uparrow$  & FPR$\downarrow$ & AUC$\uparrow$  & FPR$\downarrow$ & AUC$\uparrow$  & FPR$\downarrow$    & AUC$\uparrow$    \\
\midrule
\multicolumn{11}{l}{\textbf{CLIP-base}}                                                  \\
\multicolumn{11}{l}{\textbf{MCM Score}}                                                  \\
NPOS$^\dagger$  & 19.59 & 95.68 & 48.26 & 89.70 & 49.82 & 88.77 & 51.12 & 87.58 & 42.20 & 90.43 \\
CoOp$^\dagger$  & 43.38 & 91.26 & 38.53 & 91.95 & 46.68 & 89.09 & 50.64 & 87.83 & 44.81 & 90.03 \\
LoCoOp$^\dagger$        & 38.49 & 92.49 & 33.27 & 93.67 & 39.23 & 91.07 & 49.25 & 89.13 & 40.17 & 91.53 \\
LoCoOp$^*$      & 31.33	& 93.64	& 33.68	& 93.37	& 42.31	& 90.10	& 51.72	& 87.75	& 39.76	& 91.22 \\
\bottomrule
\end{tabular}
}
\label{tab:finetune_appendix}
\end{table}

\begin{table}[]
\caption{\textbf{Detail results of ablation study on modality.} We use CLIP-B/16 as a backbone.}
\centering
\resizebox{\textwidth}{!}{
\begin{tabular}{lcccccccccccccccc}
\toprule
\multirow{2}{*}{Method} & \multicolumn{2}{c}{iNaturalist} & \multicolumn{2}{c}{SUN} & \multicolumn{2}{c}{Places} & \multicolumn{2}{c}{Texture} & \multicolumn{2}{c}{ImageNet22K} & \multicolumn{2}{c}{COCO}         & \multicolumn{2}{c}{\textbf{Average}} \\
\cmidrule(lr){2-3} \cmidrule(lr){4-5} \cmidrule(lr){6-7} \cmidrule(lr){8-9} \cmidrule(lr){10-11} \cmidrule(lr){12-13} \cmidrule(lr){14-15} \cmidrule(lr){16-17}
                            & FPR$\downarrow$   & AUC$\uparrow$  & FPR$\downarrow$ & AUC$\uparrow$ & FPR$\downarrow$ & AUC$\uparrow$  & FPR$\downarrow$ & AUC$\uparrow$& FPR$\downarrow$  & AUC$\uparrow$ & FPR$\downarrow$  & AUC$\uparrow$ & FPR$\downarrow$      & AUC$\uparrow$    \\
\midrule
\multicolumn{17}{l}{\textbf{ImageNet1K}}                                                                                                                                                                                                                                             \\
\multicolumn{17}{l}{\textbf{MCM Score}}                                                                                                                                                                                                                                              \\
Visual                  & 29.69          & 94.58          & 35.15      & 92.99      & 41.25        & 90.45       & 55.00        & 86.92        & -              & -              & -     & -                        & 40.27             & 91.24            \\
Text                    & 30.21          & 94.33          & 38.39      & 92.27      & 44.48        & 89.74       & 58.05        & 85.64        & -              & -              & -     & -                        & 42.78             & 90.50            \\
\bottomrule
\end{tabular}
}
\label{tab:modality_appendix}
\end{table}

\begin{table}[]
    \caption{\textbf{Detail results of \mname with and without considering projection matrix $\mathrm{W_{p}}$.} We use CLIP-B/16 as a backbone.}
    \centering
    \resizebox{\textwidth}{!}{
    \begin{tabular}{lcccccccccccccccc}
    \toprule
    \multirow{2}{*}{Method} & \multicolumn{2}{c}{iNaturalist} & \multicolumn{2}{c}{SUN} & \multicolumn{2}{c}{Places} & \multicolumn{2}{c}{Texture} & \multicolumn{2}{c}{ImageNet22K} & \multicolumn{2}{c}{COCO}         & \multicolumn{2}{c}{\textbf{Average}} \\
    \cmidrule(lr){2-3} \cmidrule(lr){4-5} \cmidrule(lr){6-7} \cmidrule(lr){8-9} \cmidrule(lr){10-11} \cmidrule(lr){12-13} \cmidrule(lr){14-15} \cmidrule(lr){16-17}
                                & FPR$\downarrow$   & AUC$\uparrow$  & FPR$\downarrow$ & AUC$\uparrow$ & FPR$\downarrow$ & AUC$\uparrow$  & FPR$\downarrow$ & AUC$\uparrow$& FPR$\downarrow$  & AUC$\uparrow$ & FPR$\downarrow$  & AUC$\uparrow$ & FPR$\downarrow$      & AUC$\uparrow$    \\
    \midrule
    \multicolumn{17}{l}{\textbf{ImageNet1K}}                                                                                                                                                                                                                                     \\
    \multicolumn{17}{l}{\textbf{MCM Score}}                                                                                                                                                                                                                                      \\
    Vanilla MCM             & 32.07          & 94.43          & 38.65      & 92.37      & 43.73        & 90.03       & 57.89        & 86.13        & -          & -          & -           & -          & 43.09             & 90.74            \\
    \mname w $\mathrm{W_{p}}$   & 35.21          & 93.06          & 33.50      & 93.16      & 41.02        & 90.50       & 57.41        & 86.22        & -          & -          & -           & -          & 41.79             & 90.74            \\
\bottomrule
    \end{tabular}
    }
    \label{tab:proj_w_appendix}
\end{table}

\begin{table}[]
\caption{\textbf{Detail results for \mname with different backbones.} $^\dagger$ is cited from \citet{jiang2024negative}. $^\ast$ denotes the result of our re-run.}
\centering
\resizebox{\textwidth}{!}{
\begin{tabular}{lcccccccccccccccc}
\toprule
\multirow{2}{*}{Method} & \multicolumn{2}{c}{iNaturalist} & \multicolumn{2}{c}{SUN} & \multicolumn{2}{c}{Places} & \multicolumn{2}{c}{Texture} & \multicolumn{2}{c}{ImageNet22K} & \multicolumn{2}{c}{COCO}         & \multicolumn{2}{c}{\textbf{Average}} \\
\cmidrule(lr){2-3} \cmidrule(lr){4-5} \cmidrule(lr){6-7} \cmidrule(lr){8-9} \cmidrule(lr){10-11} \cmidrule(lr){12-13} \cmidrule(lr){14-15} \cmidrule(lr){16-17}
                            & FPR$\downarrow$   & AUC$\uparrow$  & FPR$\downarrow$ & AUC$\uparrow$ & FPR$\downarrow$ & AUC$\uparrow$  & FPR$\downarrow$ & AUC$\uparrow$& FPR$\downarrow$  & AUC$\uparrow$ & FPR$\downarrow$  & AUC$\uparrow$ & FPR$\downarrow$      & AUC$\uparrow$    \\
\midrule
\multicolumn{17}{l}{\textbf{ImageNet1K}}                                                                                                           \\
\multicolumn{17}{l}{\textbf{CLIP-base}}                                                                                                     \\
Vanilla MCM$^\ast$    & 32.07 & 94.43 & 38.65 & 92.37 & 43.73 & 90.03 & 57.89 & 86.13 & -     & -     & -     & -     & 43.09 & 90.74 \\
SeTAR+MCM       & 26.92 & 94.67 & 35.57 & 92.79 & 42.64 & 90.16 & 55.83 & 86.58 & -     & -     & -     & -     & 40.24 & 91.05 \\
Vanilla GL-MCM$^\ast$ & 15.34 & 96.62 & 30.65 & 93.01 & 37.76 & 90.07 & 57.41 & 83.73 & -     & -     & -     & -     & 35.29 & 90.86 \\
SeTAR+GL-MCM    & 13.36 & 96.92 & 28.17 & 93.36 & 36.80 & 90.40 & 54.17 & 84.59 & -     & -     & -     & -     & 33.12 & 91.32 \\
\bottomrule
\end{tabular}
}
\label{tab:backbone_appendix}
\end{table}

\begin{table}[]
\caption{\textbf{Detail results for different search algorithms.} Here LES stands for layer-exhaustive greedy search, MIS stands for modality-interleave greedy search, and SeTAR-S stands for the search algorithm of \mnamec, which searches vision and text layers sequentially. We use CLIP-B/16 as a backbone.}
\centering
\resizebox{\textwidth}{!}{
\begin{tabular}{lcccccccccccccccc}
\toprule
\multirow{2}{*}{Method} & \multicolumn{2}{c}{iNaturalist} & \multicolumn{2}{c}{SUN} & \multicolumn{2}{c}{Places} & \multicolumn{2}{c}{Texture} & \multicolumn{2}{c}{ImageNet22K} & \multicolumn{2}{c}{COCO}         & \multicolumn{2}{c}{\textbf{Average}} \\
\cmidrule(lr){2-3} \cmidrule(lr){4-5} \cmidrule(lr){6-7} \cmidrule(lr){8-9} \cmidrule(lr){10-11} \cmidrule(lr){12-13} \cmidrule(lr){14-15} \cmidrule(lr){16-17}
                            & FPR$\downarrow$   & AUC$\uparrow$  & FPR$\downarrow$ & AUC$\uparrow$ & FPR$\downarrow$ & AUC$\uparrow$  & FPR$\downarrow$ & AUC$\uparrow$  & FPR$\downarrow$ & AUC$\uparrow$ & FPR$\downarrow$  & AUC$\uparrow$ & FPR$\downarrow$      & AUC$\uparrow$    \\
\midrule
\multicolumn{17}{l}{\textbf{ImageNet1K}}                                                                                                                 \\
\multicolumn{17}{l}{\textbf{MCM Score}}                                                                                                                  \\
LES                      & 30.25 & 94.26 & 36.42 & 92.79 & 42.97 & 90.15 & 58.33 & 85.89 & -     & -     & -     & -     & 41.99 & 90.78 \\
MIS                      & 28.63 & 94.46 & 35.41 & 92.80 & 42.37 & 90.17 & 55.78 & 86.59 & -     & -     & -     & -     & 40.55 & 91.00 \\
SeTAR-S                  & 26.92 & 94.67 & 35.57 & 92.79 & 42.64 & 90.16 & 55.83 & 86.58 & -     & -     & -     & -     & 40.24 & 91.05 \\
\multicolumn{17}{l}{\textbf{GL-MCM Score}}                                                                                                               \\
\bottomrule
\end{tabular}
}
\label{tab:dif-search_appendix}
\end{table}

\begin{table}[]
    \caption{\textbf{Detail results of ablation study on $\lambda$.} We use CLIP-B/16 as a backbone.}
    \centering
    \resizebox{\textwidth}{!}{
    \begin{tabular}{lcccccccccccccccc}
    \toprule
    \multirow{2}{*}{$\lambda$} & \multicolumn{2}{c}{iNaturalist} & \multicolumn{2}{c}{SUN} & \multicolumn{2}{c}{Places} & \multicolumn{2}{c}{Texture} & \multicolumn{2}{c}{ImageNet22K} & \multicolumn{2}{c}{COCO}         & \multicolumn{2}{c}{\textbf{Average}} \\
    \cmidrule(lr){2-3} \cmidrule(lr){4-5} \cmidrule(lr){6-7} \cmidrule(lr){8-9} \cmidrule(lr){10-11} \cmidrule(lr){12-13} \cmidrule(lr){14-15} \cmidrule(lr){16-17}
                            & FPR$\downarrow$   & AUC$\uparrow$  & FPR$\downarrow$ & AUC$\uparrow$ & FPR$\downarrow$ & AUC$\uparrow$  & FPR$\downarrow$ & AUC$\uparrow$& FPR$\downarrow$   & AUC$\uparrow$ & FPR$\downarrow$  & AUC$\uparrow$ & FPR$\downarrow$      & AUC$\uparrow$    \\
    \midrule
    \multicolumn{17}{l}{\textbf{ImageNet1K}}                                                                                                                                                                                                                                     \\
    \multicolumn{17}{l}{\textbf{MCM Score}}                                                                                                                                                                                                                                      \\
    0.01                    & 28.31          & 94.60          & 36.83      & 92.74      & 43.01        & 90.10       & 55.85        & 86.58        & -              & -              & -           & -          & 41.00             & 91.00            \\
    0.05                    & 27.41          & 94.75          & 35.91      & 92.70      & 42.75        & 90.15       & 55.57        & 86.49        & -              & -              & -           & -          & 40.41             & 91.02            \\
\bottomrule
    \end{tabular}
    }
    \label{tab:lambda_appendix}
    \end{table}

\begin{table}[]
\caption{\textbf{Detail results on different pruning strategies.} We use CLIP-B/16 as a backbone.}
\centering
\resizebox{\textwidth}{!}{
\begin{tabular}{lcccccccccccccccc}
\toprule
\multirow{2}{*}{Method} & \multicolumn{2}{c}{iNaturalist} & \multicolumn{2}{c}{SUN} & \multicolumn{2}{c}{Places} & \multicolumn{2}{c}{Texture} & \multicolumn{2}{c}{ImageNet22K} & \multicolumn{2}{c}{COCO}         & \multicolumn{2}{c}{\textbf{Average}} \\
\cmidrule(lr){2-3} \cmidrule(lr){4-5} \cmidrule(lr){6-7} \cmidrule(lr){8-9} \cmidrule(lr){10-11} \cmidrule(lr){12-13} \cmidrule(lr){14-15} \cmidrule(lr){16-17}
                            & FPR$\downarrow$   & AUC$\uparrow$  & FPR$\downarrow$ & AUC$\uparrow$ & FPR$\downarrow$ & AUC$\uparrow$  & FPR$\downarrow$ & AUC$\uparrow$  & FPR$\downarrow$ & AUC$\uparrow$ & FPR$\downarrow$  & AUC$\uparrow$ & FPR$\downarrow$      & AUC$\uparrow$    \\
\midrule
\multicolumn{17}{l}{\textbf{ImageNet1K}}                                                                                                                 \\
\multicolumn{17}{l}{\textbf{MCM Score}}                                                                                                                  \\
Principle & 32.07 & 94.43 & 38.65 & 92.37 & 43.73 & 90.03 & 57.89 & 86.13 & - & - & - & - & 43.09 & 90.74 \\
Random & 32.07 & 94.43 & 38.65 & 92.37 & 43.73 & 90.03 & 57.89 & 86.13 & - & - & - & - & 43.09 & 90.74 \\
Minor & 26.92 & 94.67 & 35.57 & 92.79 & 42.64 & 90.16 & 55.83 & 86.58 & - & - & - & - & 40.24 & 91.05 \\
\multicolumn{17}{l}{\textbf{GL-MCM Score}}                                                                                                               \\
\bottomrule
\end{tabular}
}
\label{tab:prune_type_appendix}
\end{table}

\begin{table}[]
\caption{\textbf{Detail results of ablation study on top-K.} We use CLIP-B/16 as a backbone.}
\centering
\resizebox{\textwidth}{!}{
\begin{tabular}{lcccccccccccccccc}
\toprule
\multirow{2}{*}{K} & \multicolumn{2}{c}{iNaturalist} & \multicolumn{2}{c}{SUN} & \multicolumn{2}{c}{Places} & \multicolumn{2}{c}{Texture} & \multicolumn{2}{c}{ImageNet22K} & \multicolumn{2}{c}{COCO}         & \multicolumn{2}{c}{\textbf{Average}} \\
\cmidrule(lr){2-3} \cmidrule(lr){4-5} \cmidrule(lr){6-7} \cmidrule(lr){8-9} \cmidrule(lr){10-11} \cmidrule(lr){12-13} \cmidrule(lr){14-15} \cmidrule(lr){16-17}
                            & FPR$\downarrow$   & AUC$\uparrow$  & FPR$\downarrow$ & AUC$\uparrow$ & FPR$\downarrow$ & AUC$\uparrow$  & FPR$\downarrow$ & AUC$\uparrow$&  FPR$\downarrow$ & AUC$\uparrow$ & FPR$\downarrow$  & AUC$\uparrow$ & FPR$\downarrow$      & AUC$\uparrow$    \\
\midrule
\multicolumn{17}{l}{\textbf{ImageNet1K}}                                                                                                                                                                                                                                    \\
\multicolumn{17}{l}{\textbf{MCM Score}}                                                                                                                                                                                                                                     \\
0                      & 26.50          & 94.70          & 36.22      & 92.66      & 43.04        & 90.10       & 55.82        & 86.46        & -              & -              & -           & -          & 40.39             & 90.98            \\
100                    & 26.92          & 94.67          & 35.57      & 92.79      & 42.64        & 90.16       & 55.83        & 86.58        & -              & -              & -           & -          & 40.24             & 91.05            \\
\bottomrule
\end{tabular}
}
\label{tab:topk_appendix}
\end{table}

\begin{table}[]
\caption{\textbf{Detail results of ResNet50.} We use ImageNet1K as the ID dataset. $^\dagger$ is cited from \cite{djurisic2023extremely}.}
\centering
\resizebox{\textwidth}{!}{
\begin{tabular}{lcccccccccc}
\toprule
\multirow{2}{*}{Method} & \multicolumn{2}{c}{iNaturalist} & \multicolumn{2}{c}{SUN} & \multicolumn{2}{c}{Places} & \multicolumn{2}{c}{Texture} & \multicolumn{2}{c}{\textbf{Average}} \\
                        & FPR$\downarrow$           & AUC$\uparrow$           & FPR$\downarrow$       & AUC$\uparrow$       & FPR$\downarrow$         & AUC$\uparrow$        & FPR$\downarrow$         & AUC$\uparrow$         & FPR$\downarrow$              & AUC$\uparrow$             \\
\midrule
Softmax$^\dagger$                 & 54.99          & 87.74          & 70.83      & 80.86      & 73.99        & 79.76       & 68.00        & 79.61        & 66.95             & 81.99            \\
Energy$^\dagger$                  & 55.72          & 89.95          & 59.26      & 85.89      & 64.92        & 82.86       & 53.72        & 85.99        & 58.41             & 86.17            \\
ReAct$^\dagger$                   & 20.38          & 96.22          & 24.20      & 94.20      & 33.85        & 91.58       & 47.30        & 89.80        & 31.43             & 92.95            \\
DICE$^\dagger$                    & 25.63          & 94.49          & 35.15      & 90.83      & 46.49        & 87.48       & 31.72        & 90.30        & 34.75             & 90.77            \\
ASH-P$^\dagger$                   & 44.57          & 92.51          & 52.88      & 88.35      & 61.79        & 61.79       & 42.06        & 89.70        & 50.32             & 89.04            \\
ASH-B$^\dagger$                   & 14.21          & 97.32          & 22.08      & 95.10      & 33.45        & 92.31       & 21.17        & 95.50        & 22.73             & 95.06            \\
\bottomrule
\end{tabular}
}
\label{tab:resnet50_appendix}
\end{table}


\clearpage
\begin{figure}[]
	\centering
	\subfigure[MCM score]{
		\label{fig:weight_imagenet_mcm}
		\includegraphics[width=\linewidth]{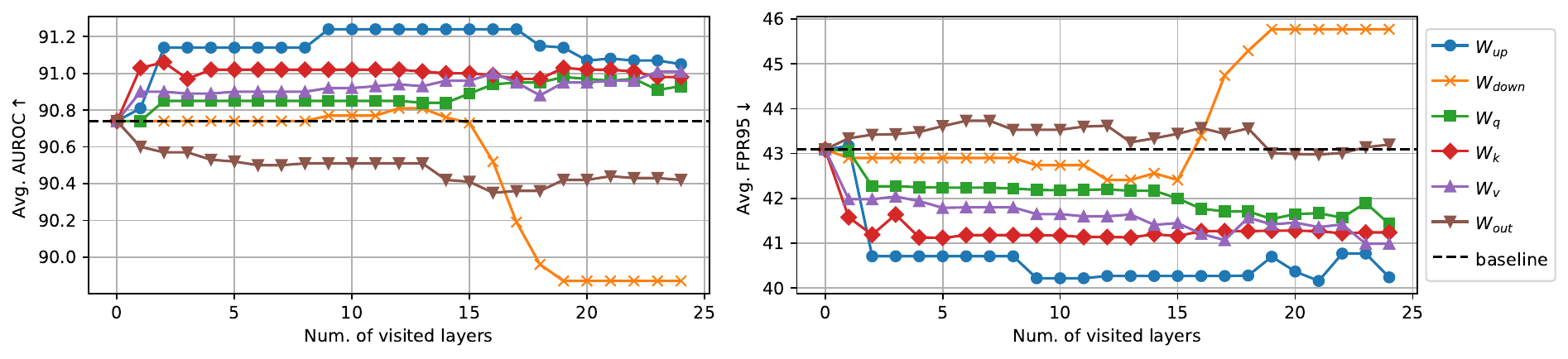}}
	\subfigure[GL-MCM score]{
		\label{fig:weight_imagenet_gl_mcm}
		\includegraphics[width=\linewidth]{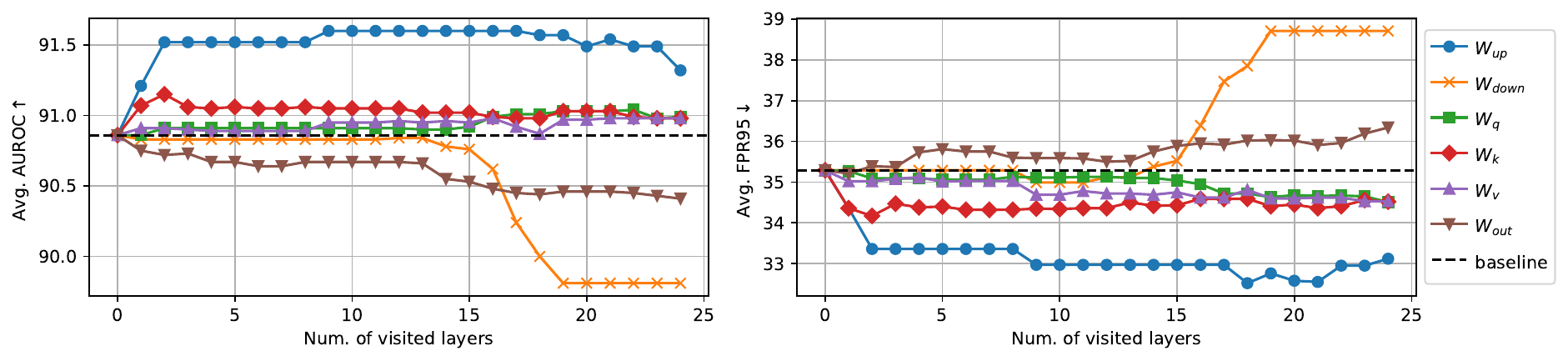}}
	\caption{\textbf{Average AUROC/FPR95 of different weight types on ImageNet1K benchmark.} We use CLIP-B/16 as a backbone.}
	\label{fig:weight_imagenet}
\end{figure}


\begin{figure}[]
	\centering
	\subfigure[MCM score]{
		\label{fig:weight_voc_mcm}
		\includegraphics[width=\linewidth]{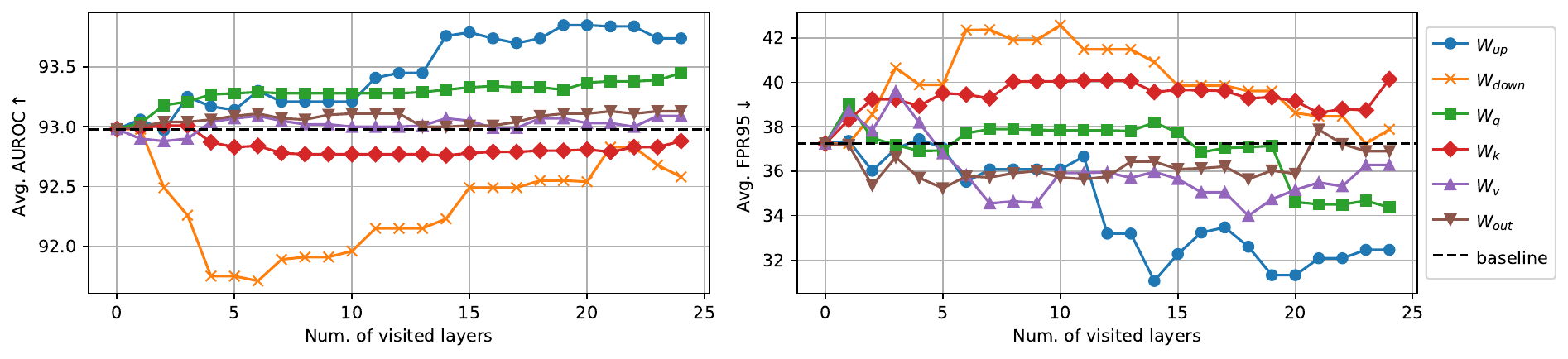}}
	\subfigure[GL-MCM score]{
		\label{fig:weight_voc_gl_mcm}
		\includegraphics[width=\linewidth]{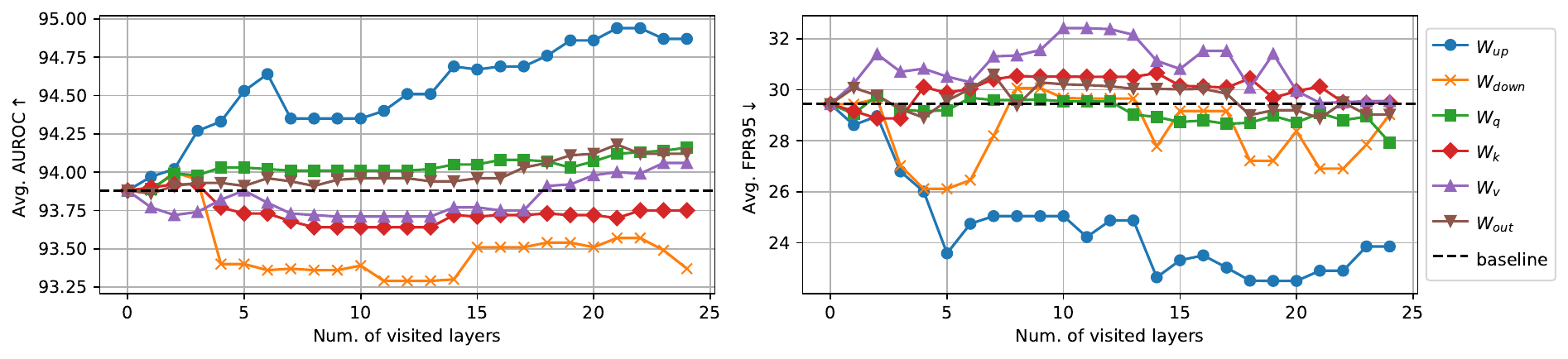}}
	\caption{\textbf{Average AUROC/FPR95 of different weight types on Pascal-VOC benchmark.} We use CLIP-B/16 as a backbone.}
	\label{fig:weight_voc}
\end{figure}

\begin{figure}[]
	\centering
	\subfigure[ImageNet1K]{
		\label{fig:lambda_imagenet}
		\includegraphics[width=\linewidth]{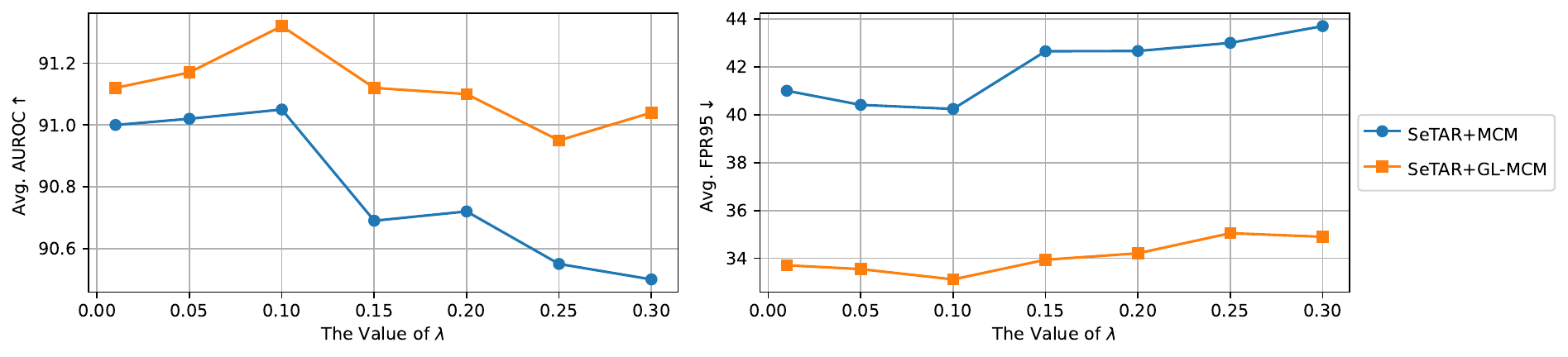}}
	\subfigure[Pascal-VOC]{
		\label{fig:lambda_voc}
		\includegraphics[width=\linewidth]{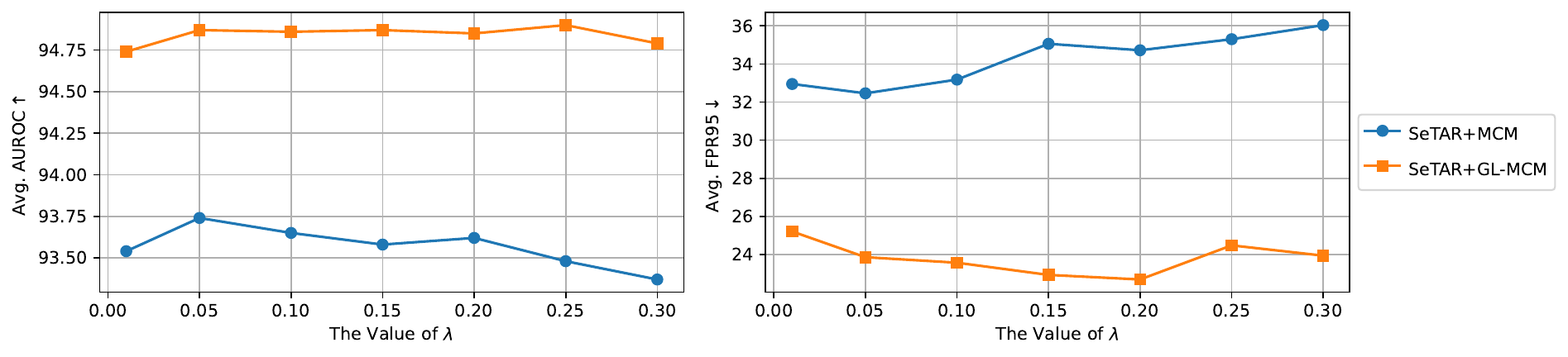}}
	\caption{\textbf{Ablation studies on $\lambda$ on different ID datasets.} We use CLIP-B/16 as a backbone.}
	\label{fig:appendix_lambda_abs}
\end{figure}

\begin{figure}[]
	\centering
	\subfigure[ImageNet1K]{
		\label{fig:topk_imagenet}
		\includegraphics[width=\linewidth]{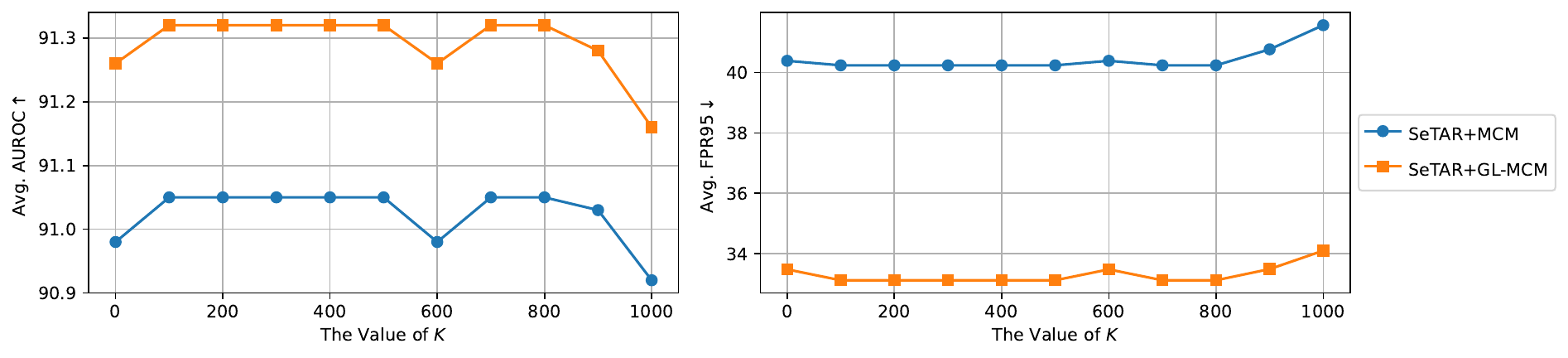}}
	\subfigure[Pascal-VOC]{
		\label{fig:topk_voc}
		\includegraphics[width=\linewidth]{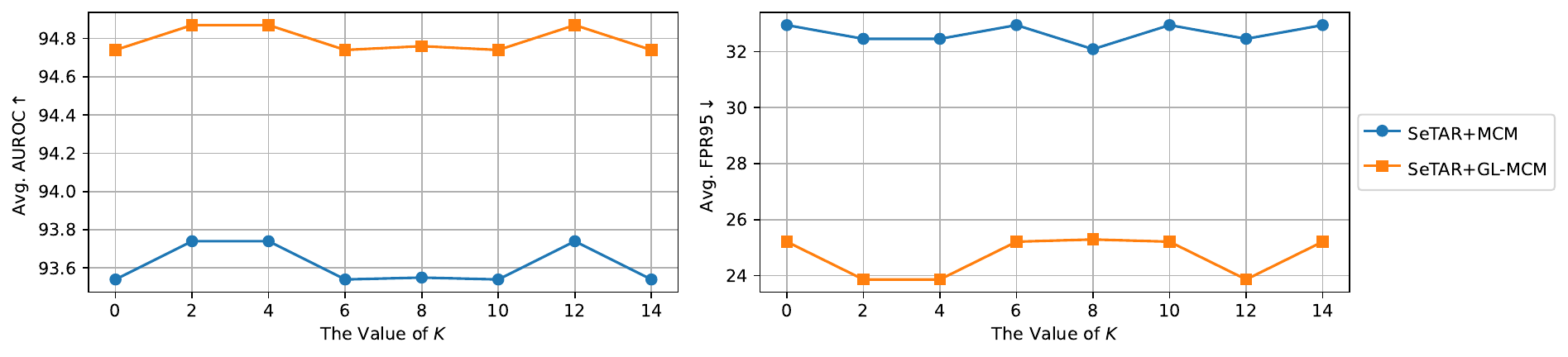}}
	\caption{\textbf{Ablation studies on top-K on different ID datasets.} We use CLIP-B/16 as a backbone.}
	\label{fig:appendix_topk_abs}
\end{figure}

\begin{figure}[]
	\centering
	\subfigure[LoCoOp Loss]{
		\label{fig:total_loss}
		\includegraphics[width=0.32\linewidth]{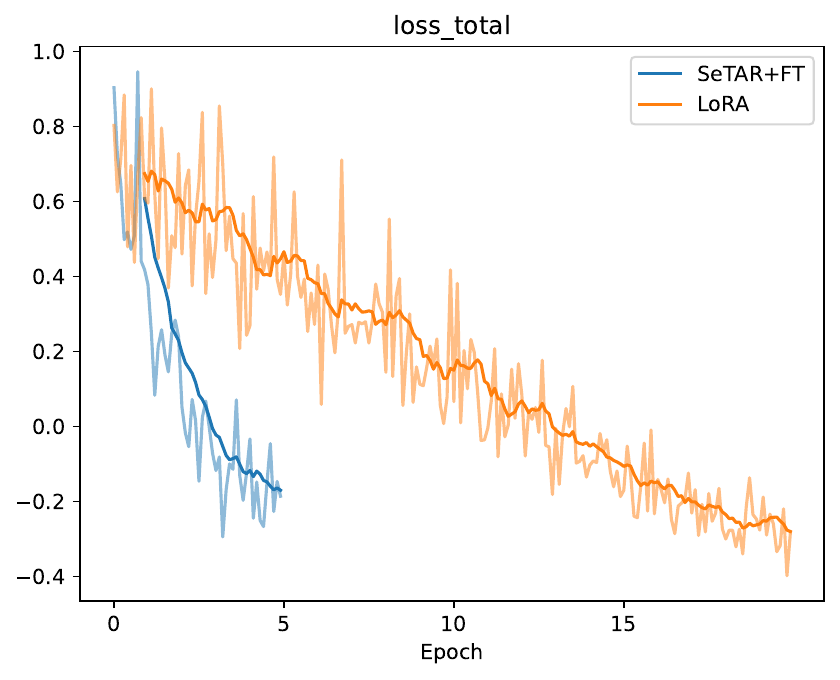}}
	\subfigure[ID Loss]{
		\label{fig:id_loss}
		\includegraphics[width=0.32\linewidth]{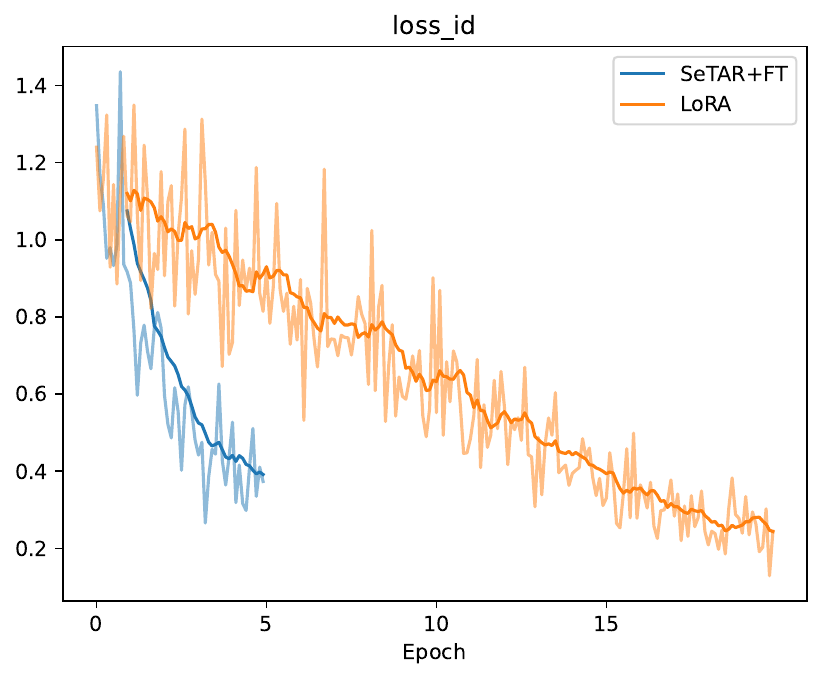}}
	\subfigure[OOD Loss]{
		\label{fig:ood_loss}
		\includegraphics[width=0.32\linewidth]{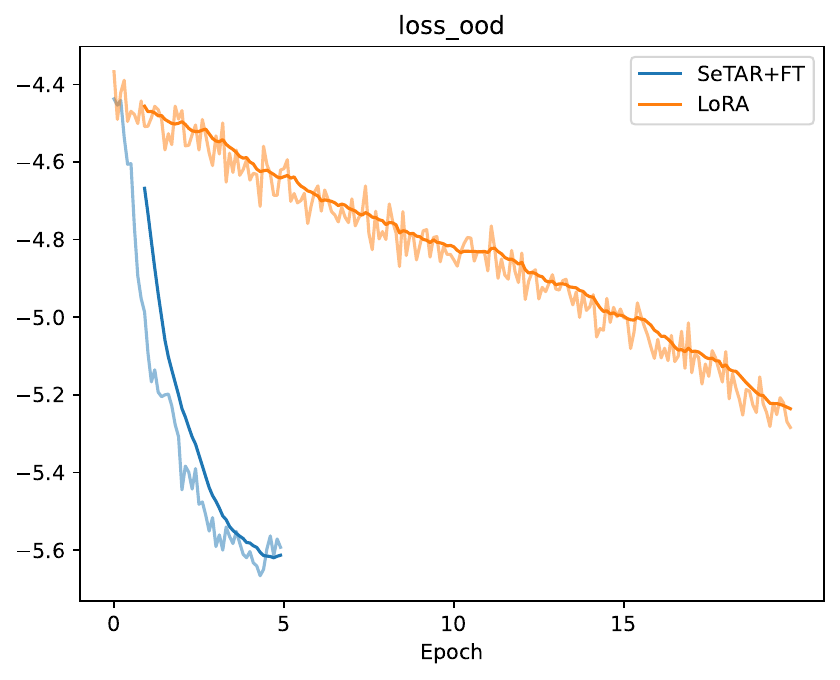}}
	\caption{\textbf{Loss plots of \mnamec+FT v.s. LoRA on ImageNet1K.} We use CLIP-B/16 as a backbone. \mnamec+FT demonstrates faster convergence across all losses, especially in the OOD loss. For reference, with MCM score, \mnamec+FT achieves an average FPR of 38.77 at epoch 5. While LoRA achieves an average FPR of 42.88, 39.92 and 42.23 at epoch 1, 5 and 15, respectively.}
	\label{fig:loss_plot}
\end{figure}\begin{figure}[]
	\centering
	\subfigure[CLIP-base]{
		\label{fig:rank_ratio_clip-base}
		\includegraphics[width=\linewidth]{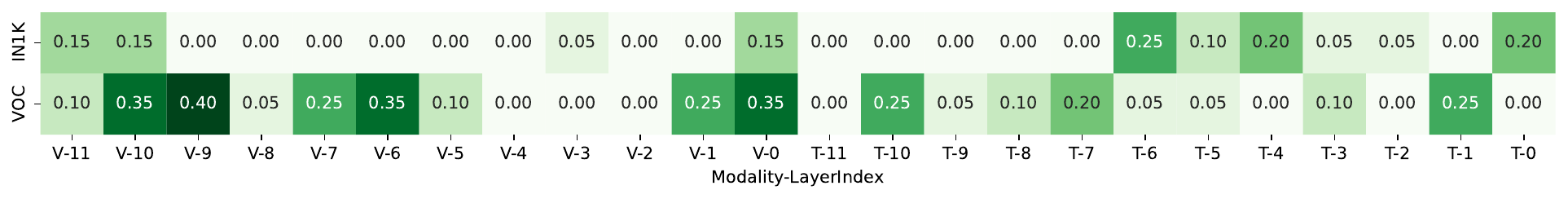}}
	\subfigure[CLIP-large]{
		\label{fig:rank_ratio_clip-large}
		\includegraphics[width=\linewidth]{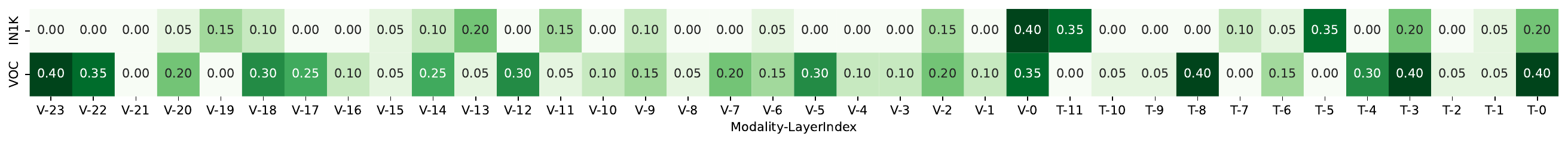}}
	\subfigure[Swin-base]{
		\label{fig:rank_ratio_swin-base}
		\includegraphics[width=\linewidth]{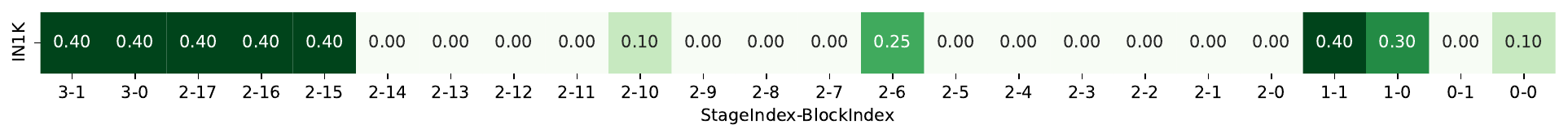}}
    \caption{\textbf{Visualization of \mname rank reduction ratio distribution on different ID datasets with different backbones.} IN1K, VOC stand for ImageNet1K and Pascal-VOC. And V, T stand for visual modality and text modality of the CLIP model.}
	\label{fig:ratio_plot}
\end{figure}

\clearpage

\begin{lstlisting}[caption={\textbf{Example procedure of \mname on ImageNet1K with CLIP-base.} We search the visual and text tower from top to bottom. At each step, we select the best ratio that minimizes the loss.}, label={lst:log}]
     tower_type weight_type  layer_num  best_ratio   total_loss*   id_loss  ood_loss   val_acc   ood_patch_percent
step
0        visual        W_up         11        0.15      0.647777  1.093326 -4.455494  71.399998          38.906631
1        visual        W_up         10        0.15      0.644654  1.083629 -4.389751  71.799998          39.293876
2        visual        W_up          9        0.00      0.644654  1.083629 -4.389751  71.799998          39.293876
3        visual        W_up          8        0.00      0.644654  1.083629 -4.389751  71.799998          39.293876
4        visual        W_up          7        0.00      0.644654  1.083629 -4.389751  71.799998          39.293876
5        visual        W_up          6        0.00      0.644654  1.083629 -4.389751  71.799998          39.293876
6        visual        W_up          5        0.00      0.644654  1.083629 -4.389751  71.799998          39.293876
7        visual        W_up          4        0.00      0.644654  1.083629 -4.389751  71.799998          39.293876
8        visual        W_up          3        0.05      0.640844  1.079729 -4.388844  71.999998          39.209695
9        visual        W_up          2        0.00      0.640844  1.079729 -4.388844  71.999998          39.209695
10       visual        W_up          1        0.00      0.640844  1.079729 -4.388844  71.999998          39.209695
11       visual        W_up          0        0.15      0.640132  1.079109 -4.389775  72.199998          39.156123
12         text        W_up         11        0.00      0.640132  1.079109 -4.389775  72.199998          39.156123
13         text        W_up         10        0.00      0.640132  1.079109 -4.389775  72.199998          39.156123
14         text        W_up          9        0.00      0.640132  1.079109 -4.389775  72.199998          39.156123
15         text        W_up          8        0.00      0.640132  1.079109 -4.389775  72.199998          39.156123
16         text        W_up          7        0.00      0.640132  1.079109 -4.389775  72.199998          39.156123
17         text        W_up          6        0.25      0.630751  1.075123 -4.443716  71.600001          38.808673
18         text        W_up          5        0.10      0.630514  1.078703 -4.481889  71.599997          38.246428
19         text        W_up          4        0.20      0.622065  1.075958 -4.538932  72.000001          38.452552
20         text        W_up          3        0.05      0.620440  1.079326 -4.588857  71.999997          38.649488
21         text        W_up          2        0.05      0.618521  1.076858 -4.583368  71.600001          38.444899
22         text        W_up          1        0.00      0.618521  1.076858 -4.583368  71.600001          38.444899
23         text        W_up          0        0.20      0.615174  1.069851 -4.546776  72.499997          38.642345
\end{lstlisting}

\end{document}